\pdfoutput=1
\documentclass[10pt,twocolumn,letterpaper]{article}

\usepackage{cvpr}
\usepackage{times}
\usepackage{epsfig}
\usepackage{graphicx}
\usepackage{amsmath}
\usepackage{amssymb}
\usepackage{subcaption}
\usepackage{caption}
\usepackage{xcolor}
\usepackage[font=small,labelfont=bf,tableposition=top]{caption}
\usepackage{booktabs}
\usepackage{multirow}
\usepackage{algorithm,algorithmic}
\usepackage[numbers,sort]{natbib} 
\usepackage{authblk}
\usepackage{xspace}
\usepackage{enumitem}
\usepackage{pifont}

\usepackage{hyperref}
\hypersetup{pagebackref=true,breaklinks=true,colorlinks,bookmarks=false}

\cvprfinalcopy

\newcommand{\xmark}{\ding{55}}

\newcommand{\blockb}[3]{
  \(\left[
      \begin{array}{c}
        \text{1$\times$1, #1}\\[-.1em]
        \text{3$\times$3, #1}\\[-.1em]
        \text{1$\times$1, #2}
      \end{array}
    \right]\)$\times$#3
}

\newcommand{\blockd}[3]{
  \(\left[
      \begin{array}{c}
        \text{#1}\\[-.1em]
        \text{#2}\\[-.1em]
        \text{#3}\\
      \end{array}
    \right]\)
}

\ifcvprfinal\fi
\begin{document}
\newcommand{\repnet}{RepNet\xspace}
\newcommand{\countix}{Countix\xspace}
\newcommand{\tr}{^{^\intercal}} 

\makeatletter
\renewcommand\AB@affilsepx{  \protect\Affilfont}
\renewcommand\thesubfigure{\thefigure\alph{subfigure}} 
\renewcommand\p@subfigure{}
\makeatother

\title{
Counting Out Time: Class Agnostic Video Repetition Counting in the Wild
}

\author[ 1]{Debidatta Dwibedi}
\author[ 2]{Yusuf Aytar}
\author[ 1]{Jonathan Tompson}
\author[ 1]{Pierre Sermanet}
\author[ 2]{Andrew Zisserman}
\affil[ 1 ]{Google Research}\affil[ 2 ]
{DeepMind\protect\\\tt\small \{debidatta, yusufaytar, tompson, sermanet, zisserman\}@google.com}

\maketitle

\begin{abstract}
We present an approach for estimating the period with which an action is repeated in a video. The crux of the approach lies in constraining the period prediction module to use temporal self-similarity as an intermediate representation bottleneck that allows generalization to unseen repetitions in videos in the wild. We train this model, called \repnet, with a synthetic dataset that is generated from a large unlabeled video collection by sampling short clips of varying lengths and repeating them with different periods and counts. This combination of synthetic data and a powerful yet constrained model, allows us to predict periods in a class-agnostic fashion. Our model substantially exceeds the state of the art performance on existing periodicity (PERTUBE) and repetition counting (QUVA) benchmarks. We also collect a new challenging dataset called \countix ($\sim$90 times larger than existing datasets) which captures the challenges of repetition counting in real-world videos. Project webpage: \url{https://sites.google.com/view/repnet}.
\end{abstract}

\section{Introduction}

\begin{figure}
  \includegraphics[width=0.48\textwidth]{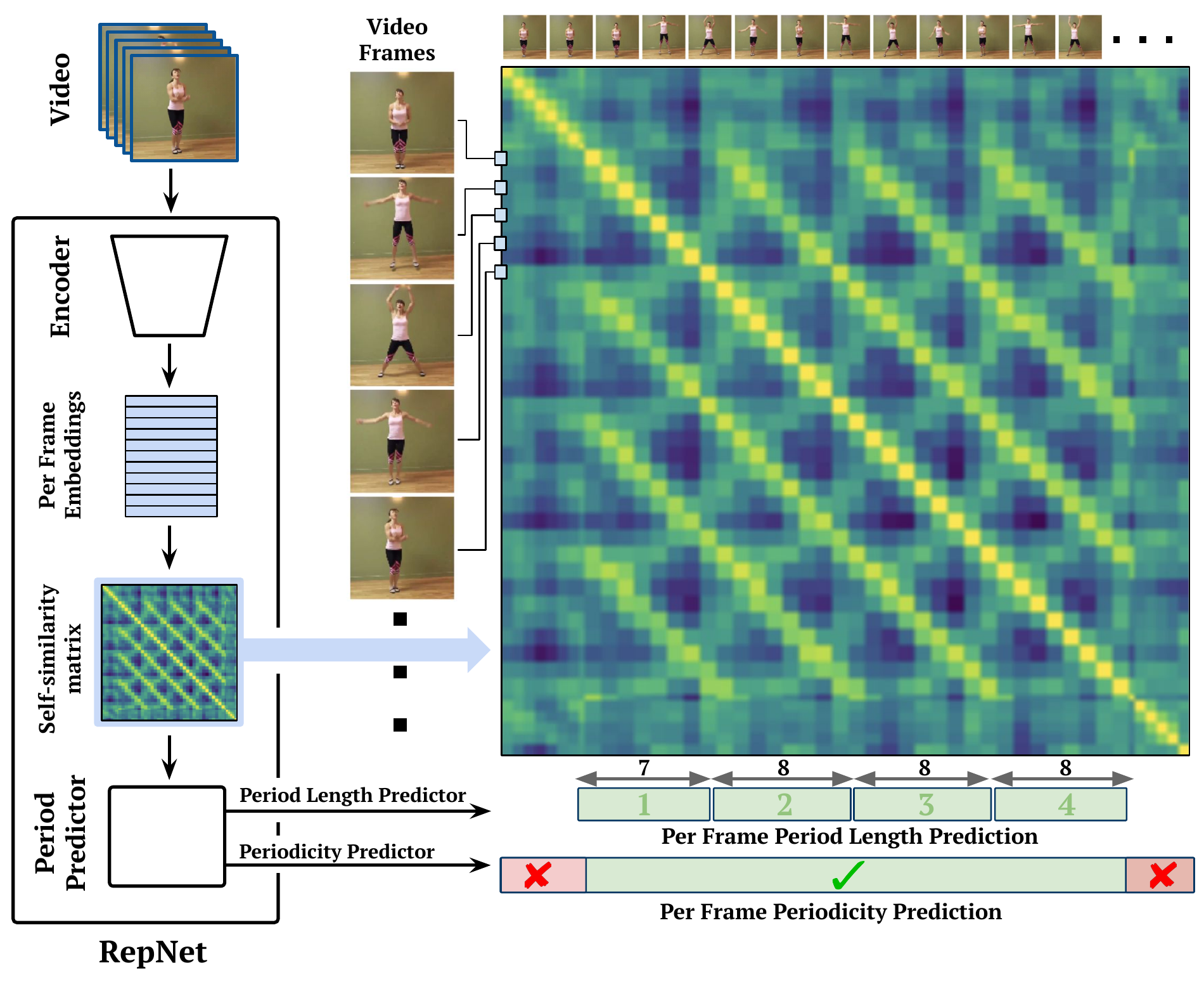}
  \caption{We present \repnet, which leverages a temporal self-similarity matrix as an intermediate layer to predict the period length and periodicity of each frame in the video.}
  \label{fig:teaser}
\end{figure}

Picture the most mundane of scenes -- a person eating by themselves in a cafe. They might be stirring sugar in their coffee while chewing their food, and tapping their feet to the background music. This person is doing at least three periodic activities in parallel. Repeating actions and processes are ubiquitous in our daily lives. These range from organic cycles, such as heart beats and breathing, through programming and manufacturing, to planetary cycles like the day-night cycle and seasons. 
Thus the need for recognizing repetitions in videos is pervasive, and a system that is able to identify and count repetitions in video will benefit any perceptual system that aims to observe and understand our world for an extended period of time.

Repetitions are also interesting for the following reasons: (1) there is usually an intent or a driving cause behind something happening multiple times; (2) the same event can be observed again but with slight variations; (3) there may be gradual changes in the scene as a result of these repetitions; (4) they provide us with unambiguous \textit{action units}, a sub-sequence in the action that can be segmented in time (for example if you are chopping an onion, the action unit is the manipulation action that is repeated to produce additional slices). Due to the above reasons, any agent interacting with the world would benefit greatly from such a system. Furthermore, repetition counting is pertinent for many computer vision applications; such as counting the number of times an exercise was done, measurement of biological events (like heartbeats), etc. 

Yet research in periodic video understanding has been limited, potentially due to the lack of a large scale labeled video repetition dataset. In contrast, for action recognition there are large scale datasets, like Kinetics~\cite{kay2017kinetics}, but their collection at large scale is enabled by the availability of keywords/text associated with the videos. Unfortunately it is rare for videos to be labeled with annotations related to repeated activity as the text is more likely to describe the semantic content. For this reason, we use a dataset with semantic action labels typically used for action recognition (\textit{Kinetics}) and manually choose videos of those classes with periodic motion (\textit{bouncing, clapping} etc.). We proceed to label the selected videos with the number of repetitions present in each clip.

Manual labelling limits the number of videos that can be annotated -- labelling is tedious and expensive due to the temporally fine-grained nature of the task. In order to increase the amount of training data, we propose a method to create synthetic repetition videos by repeating clips from existing videos with different periods. Since we are synthesizing these videos, we also have precise annotations for the period and count of repetitions in the videos, which can be used for training models using supervised learning. However, as we find in our work, such synthetic videos fail to capture all the nuances of real repeated videos and are prone to over-fitting by high-capacity deep learning models. To address this issue, we propose a data augmentation strategy for synthetic videos so that models trained on them transfer to real videos with repetitions. We use a combination of real and synthetic data to develop our model.

In this paper, our objective is a single model that works for many classes of periodic videos, and indeed, also for classes of videos unseen during training. We achieve this by using an intermediate representation that encourages generalization to unseen classes. This representation -- a temporal self-similarity matrix -- is used to predict the period with which an action is repeating in the video. This common representation is used across different kinds of repeating videos enabling the desired generalization. For example, whether a person is doing push ups, or a kid is swinging in a playground, the self-similarity matrix is the shared parameterization from which the number of repetitions is inferred. This extreme bottleneck (the number of channels in the feature map reduces from 512 to 1) also aids generalization from synthetic data to real data. The other advantage of this representation is that model interpretability is baked into the network architecture as we force the network to predict the period from the self-similarity matrix only, as opposed to inferring the period from latent high-dimensional features. 

We focus on two tasks: (i) {\em Repetition counting}, identifying the number of repeats in the video. We rephrase this problem as first estimating per frame period lengths, and then converting them to a repetition count; (ii) {\em Periodicity detection}, identifying if the current frame is a part of a repeating temporal pattern or not. We approach this as a per-frame binary classification problem.  A visual explanation of these tasks and the overview of our solution is shown in Figure~\ref{fig:teaser}.

Our main contributions in this paper are: (i) \repnet, a neural network architecture designed for counting repetitions in videos in the wild. (ii) A method to generate and augment synthetic repetition videos from unlabeled videos. (iii) By training \repnet on the synthetic dataset we outperform the state-of-the-art methods on both repetition counting and periodicity detection tasks over existing benchmarks by a substantial margin. (iv) A new video repetition counting dataset, \textit{\countix}, which is $\sim$ 90 times larger than the previous largest dataset.

\begin{figure*}
  \includegraphics[width=\textwidth]{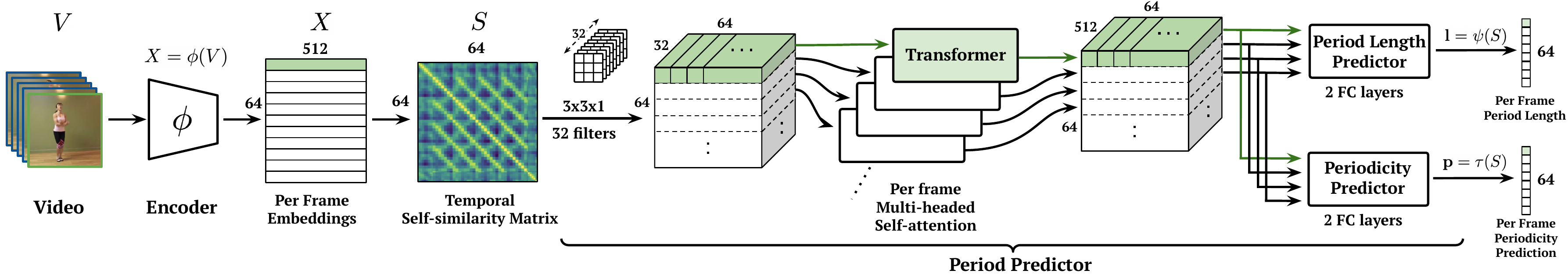}
  \caption{\textbf{\repnet architecture.} The features  produced by a single video frame is highlighted with the green color throughout the network.}
  \label{fig:repnet_architecture}
\end{figure*}

\section{Related Work}

\noindent\textbf{Self-similarity.} 
The idea of using local image and spatio-temporal self-similarities was explored in~\cite{SelfSim_ShechtmanIrani07} for pattern matching in images and videos. Matching the abstraction of self-similarities, rather than image features directly, enabled generalization. We build on this insight in our work.

\noindent\textbf{Periodicity Estimation.} Extracting periodicity (detection of periodic motion) and the period by leveraging the auto-correlation in time series is a well-studied problem~\cite{vlachos2005periodicity, stoica2005spectral}. 
 
Period estimation in videos has been done using periodograms on top of auto-correlation~\cite{cutler2000robust} or Wavelet transforms on hand-designed features derived from optical flow~\cite{runia2018real}. 
The extracted periodic motion has supported multiple tasks including 3D reconstruction~\cite{belongie2004structure, li2018structure} and bird species classification~\cite{li2013automatic}. 
Periodicity has been used  for various  applications~\cite{cutler2000robust, seitz1997view, niyogi1994analyzing, pintea2018hand} including temporal pattern classification~\cite{pogalin2008visual}.

\noindent\textbf{Temporal Self-similarity Matrix (TSM).} TSMs are useful representations for human action recognition~\cite{junejo2010view,sun2015exploring,korner2013temporal} and gait analysis~\cite{benabdelkader2004gait,benabdelkader2001eigengait} due to their robustness against large viewpoint changes when paired with appropriate feature representations. A TSM based on Improved Dense Trajectories~\cite{wang2013action} is used 
in~\cite{panagiotakis2018unsupervised} for unsupervised  identification of periodic segments in videos using special filters. Unlike these approaches, we use TSM as an intermediate layer in an end-to-end neural network architecture, which acts as an information bottleneck. Concurrently, \cite{karvounas2019reactnet} have proposed a convolutional architecture for periodicity detection in videos. 

\noindent\textbf{Synthetic Training Data.} The use of synthetic training data in computer vision is becoming more common place. Pasting object patches on real images has been shown to be effective as training data for object detection~\cite{dwibedi2017cut,tremblay2018training,georgakis2017synthesizing} and human pose estimation~\cite{Varol_2017_CVPR}. Blending multiple videos or multiple images together has been useful for producing synthetic training data for specific tasks~\cite{alayrac2019visual} as well as regularizing deep learning models~\cite{zhang2017mixup,yun2019cutmix}.
Synthetic data for training repetition counting was first proposed by~\cite{levy2015live}. They introduce a dataset of synthetic repeating patterns and use this to train a deep learning based counting model. However, the data they use for training consists of hand-designed random patterns that do not appear realistic. As shown in \cite{runia2018real}, these patterns are not diverse enough to capture all the nuances of repetitions in real videos. Instead, we propose to create synthetic training dataset of realistic video repetitions from existing video datasets.

\noindent\textbf{Counting in Computer Vision.} Counting objects and people in images~\cite{lempitsky2010learning,arteta2016counting,boominathan2016crowdnet,lu2018class,xie2018microscopy} is an active area in computer vision. On the other hand, video repetition counting~\cite{levy2015live, runia2018real} has attracted less attention from the community in the deep learning era. We build on the idea of~\cite{levy2015live} of predicting the period (cycle length), though~\cite{levy2015live} did not use a TSM.

\noindent\textbf{Temporally Fine-grained Tasks.} Repetition counting and periodicity detection are temporally fine-grained tasks like temporal action localization~\cite{shou2016temporal,chao2018rethinking}, per-frame phase classification~\cite{dwibedi2019temporal} and future anticipation~\cite{damen2018scaling}. We leverage the interfaces previously used to collect action localization datasets such as \cite{sigurdsson2016hollywood,kuehne2014language,gu2018ava} to create our repetition dataset \countix. Instead of annotating semantic segments, we label the extent of the periodic segments in videos and the number of repetitions in each segment. 
\section{\repnet Model}

In this section we introduce our \repnet architecture, which is composed of two learned components, the encoder and the period predictor, with a temporal self-similarity layer in between them. 

Assume we are given a video $V=[v_1, v_2, ... , v_N]$ as a sequence of $N$ frames. First we feed the video $V$ to an image encoder $\phi$ as $X=\phi(V)$ to produce per-frame embeddings $X=[x_1, x_2, ... , x_N]\tr$. Then, using the embeddings $X$ we obtain the self-similarity matrix $S$ by computing pairwise similarities $S_{ij}$ between all pairs of embeddings.

Finally, $S$ is fed to the period predictor module which outputs two elements for each frame: period length estimate $l=\psi(S)$ and periodicity score $p=\tau(S)$. The period length is the rate at which a repetition is occurring while the periodicity score indicates if the frame is within a periodic portion of the video or not. The overall architecture can be viewed in the Figure~\ref{fig:teaser} and a more detailed version can be seen in Figure~\ref{fig:repnet_architecture}.

\subsection{Encoder}

Our encoder $\phi$ is composed of three main components:

\noindent\textbf{Convolutional feature extractor:} We use ResNet-50\cite{he2016deep} architecture as our base 
convolutional neural network (CNN) to extract 2D convolutional features from individual frames $v_i$ of the input video. These frames are  $112\times112\times3$ in size. We use the output of \verb|conv4_block3| layer to have a larger spatial 2D feature map. The resulting per-frame features are of size $7\times7\times1024$.

\noindent\textbf{Temporal Context:} We pass these convolutional features through a layer of 3D convolutions to add local temporal information to the per-frame features. We use $512$ filters of size $3\times3\times3$ with ReLU activation with a dilation rate of 3. The temporal context helps modeling short-term motion~\cite{dwibedi2018learning, xie2018rethinking} and enables the model to distinguish between similar looking frames but with different motion (e.g.\ hands moving up or down while exercising). 

\noindent\textbf{Dimensionality reduction:} We reduce the dimensionality of extracted spatio-temporal features by using Global 2D Max-pooling over the spatial dimensions and to produce embedding vectors $x_i$ corresponding to each frame $v_i$ in the video. By collapsing the spatial dimensions we remove the need for tracking the region of interest as done explicitly in prior methods~\cite{cutler2000robust,benabdelkader2004gait,pogalin2008visual}.

\subsection{Temporal Self-similarity Matrix (TSM)}

After obtaining latent embeddings $x_i$ for each frame $v_i$, we construct the self-similarity matrix $S$ by computing all pairwise similarities $S_{ij} = f(x_i,x_j)$ between pairs of embeddings $x_i$ and $x_j$, where $f(.)$ is the similarity function. We use the negative of the squared euclidean distance as the similarity function, $f(a, b)= - ||a-b||^2$, followed by row-wise softmax operation.

\begin{figure}
  \includegraphics[width=0.48\textwidth]{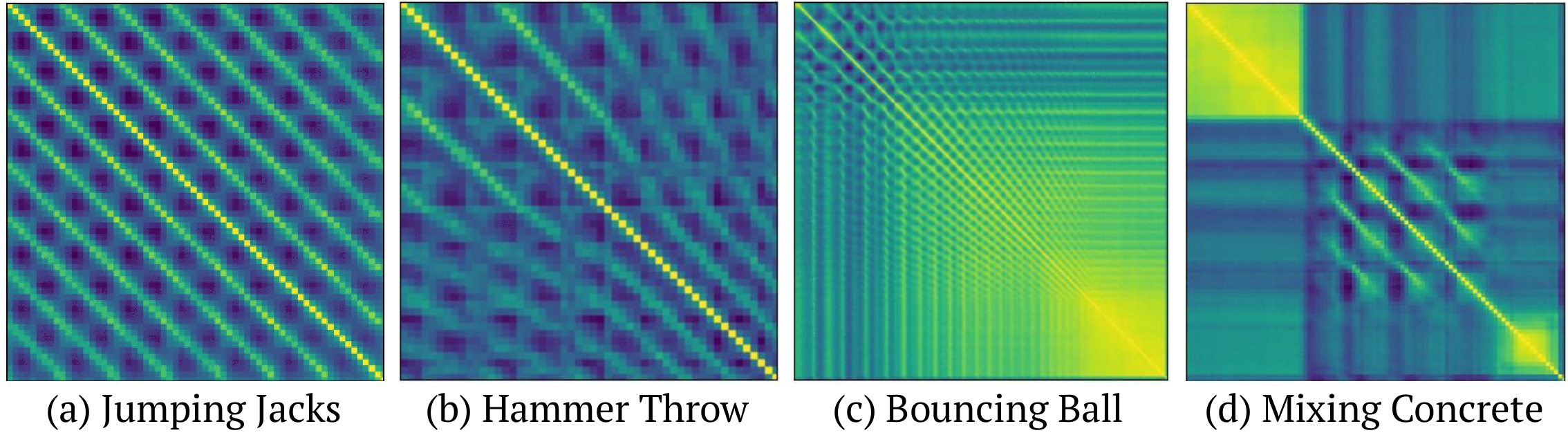}
  \caption{Diversity of temporal self-similarity matrices found in real-world repetition videos (yellow means high similarity, blue means low similarity). (a) Uniformly repeated periodic motion (jumping jacks) (b) Repetitions with acceleration (athlete performing hammer throw) (c) Repetitions with decreasing period (a bouncing ball losing speed due to repeated bounces) (d) Repeated motion preceded and succeeded by no motion (waiting to mix concrete, mixing concrete, stopped mixing). A complex model is needed to predict the period and periodicity from such diverse self-similarity matrices.}
  \label{fig:self_similarity_matrices}
\end{figure}

As the TSM has only one channel, it acts as an information bottleneck in the middle of our network and provides regularization. TSMs also make the model \textit{temporally} interpretable which brings further insights to the predictions made by the model. Some examples can be viewed in Figure~\ref{fig:self_similarity_matrices}.

\subsection{Period Predictor}

The final module of \repnet is the period predictor. This module accepts the self-similarity matrix $S=[s_1, s_2, ... , s_N]\tr$ where each row $s_i$ is the per frame self-similarity representation, and generates two outputs: per frame period length estimation $l=\psi(S)$, and per-frame binary periodicity classification $p=\tau(S)$. Note that both $l$ and $p$ are vectors and their elements are per frame predictions (i.e.\ $l_i$ is the predicted period length for the $i^{th}$ frame).

The architecture of the period predictor module can be viewed in Figure~\ref{fig:repnet_architecture}. Note that predictors $\psi$ and $\tau$ share a common architecture and weights until the last classification phase. The shared processing pipeline starts with $32$ 2D convolutional filters of size $3\times3$, followed by a transformer~\cite{vaswani2017attention} layer which uses a multi-headed attention with trainable positional embeddings in the form of a 64 length variable that is learned by training. We use 4 heads with 512 dimensions in the transformer with each head being 128 dimensions in size. After the shared pipeline, we have two classifiers, period length classifier $\psi$ and periodicity classifier $\tau$. Each of them consists of two fully connected layers of size 512.

\subsection{Losses}

Our periodicity classifier $\tau$ outputs per frame periodicity classification $p_i$ and uses a binary classification loss (binary cross-entropy) for optimization.
Our period length estimator $\psi$ outputs per frame period length estimation $l_i \in L$ where the classes are discrete period lengths $L=\{2,3,...,\frac{N}{2}\}$ where $N$ is the number of input frames. We use a multi-class classification objective (softmax cross-entropy) for optimizing our model. For all our experiments we use $N=64$. We sample the input video with different frame rates as described below to predict larger period lengths.

\subsection{Inference}
Inferring the count of repetitions robustly for a given video requires two main operations:

\noindent\textbf{Count from period length predictions:}
We sample consecutive non-overlapping windows of $N$ frames and provide it as input to \repnet which outputs per-frame periodicity $p_i$ and period lengths $l_i$.
We define \textit{per-frame count} as $\frac{p_i}{l_i}$.
The overall repetition count is computed as the sum of all per-frame counts: $\sum_{i=1}^{N} \frac{p_i}{l_i}$. The evaluation datasets for repetition counting have only periodic segments. Hence, we set $p_i$ to $1$ as default for counting experiments. 

\noindent\textbf{Multi-speed evaluation}: As our model can predict period lengths up to $32$, for covering much longer period lengths we sample input video with different frame rates. (i.e.\ we play the video at $1\times$, $2\times$, $3\times$, and $4\times$ speeds). We choose the frame rate which has the highest score for the predicted period. This is similar to what \cite{levy2015live} do at test time. 

\begin{figure}
  \includegraphics[width=0.5\textwidth]{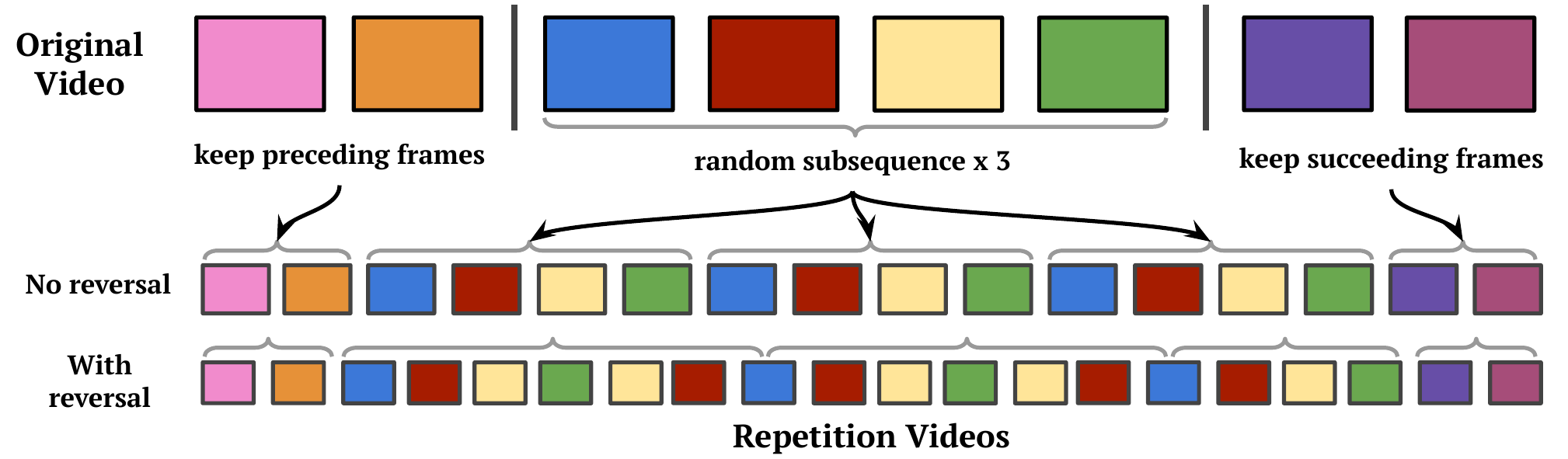}
  \caption{Our synthetic data generation pipeline that produces videos with repetitions from any video. We randomly sample a portion of a video that we repeat N times to produce synthetic repeating videos. More details in Section~\ref{sec:synthetic_dataset}}
  \label{fig:pipeline}
    \vspace{-1em}
\end{figure}

\section{Training with Synthetic Repetitions}
\label{sec:synthetic_dataset}
 
A potential supervised approach to period estimation would be collecting a large training set of periodic videos and annotating the beginning and the end of every period in all repeating actions. However, collecting such a dataset is expensive due to the fine-grained nature of the task. 

As a cheaper and more scalable alternative, we propose a training strategy that makes use of synthetically generated repetitions using unlabeled videos in the wild (e.g.\ YouTube). We generate synthetic periodic videos using randomly selected videos, and predict per frame periodicity and period lengths.
Next, we'll explain how we generate synthetic repetitions, and introduce camera motion augmentations which are crucial for training effective counting models from synthetic videos.

\subsection{Synthetic Repetition Videos}

Given a large set of unlabeled videos, we propose a simple yet effective approach for creating synthetic repetition videos (shown in Figure \ref{fig:pipeline}) from them. The advantage of using real videos to create synthetic data is that the training data is much closer to real repeated videos when compared to using synthetic patterns. Another advantage of using real videos is that using a big dataset like Kinetics ensures that the diversity of data seen by the model is huge. This allows us to train big complex models that can work on real repetition videos.

Our pipeline starts with sampling a random video \textit{V} from a dataset of videos. We use the training set of Kinetics \cite{kay2017kinetics} \textbf{without any labels}. Then, we sample a clip \textit{C} of random length \textit{P} frames from \textit{V}. This clip \textit{C} is repeated \textit{K} times (where $K > 1$) to simulate videos with repetitions. We randomly concatenate the reversed clip before repeating to simulate actions where the motion is done in reverse in the period (like jumping jacks). Then, we pre-pend and append the repeating frames with other non-repeating segments from $V$, which are just before and after $C$, respectively. The lengths of these aperiodic segments are chosen randomly and can potentially be zero too. 

This operation makes sure that there are both periodic and non-periodic segments in the generated video. Finally, each frame in the repeating part of the generated video is assigned a period length label \textit{P}. A periodicity label is also generated indicating whether the frame is inside or outside the repeating portion of the generated video.

\begin{figure}
  \includegraphics[height=5cm,keepaspectratio]{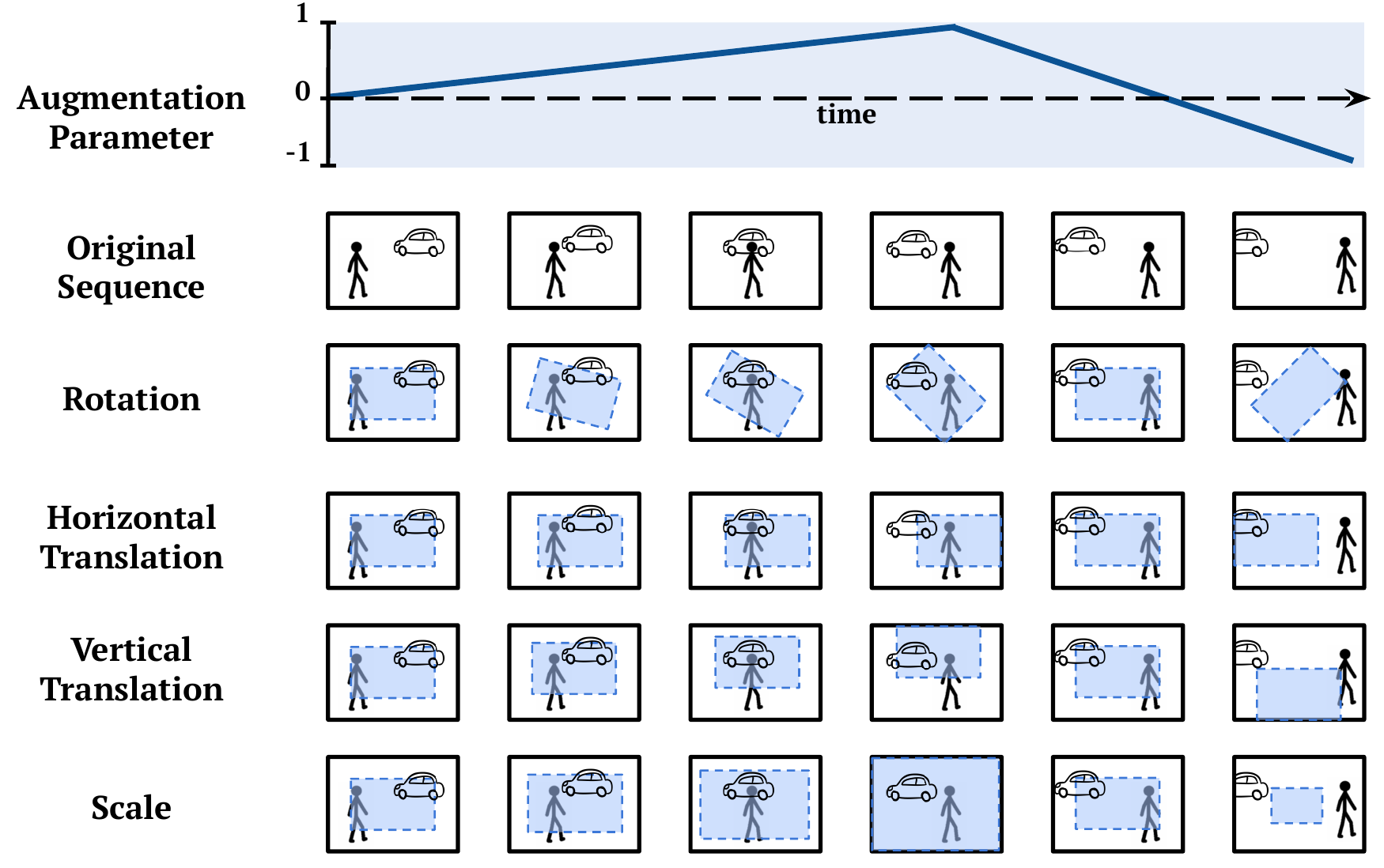}
  \caption{\textbf{Camera motion augmentation}. We vary the augmentation parameters for each type of camera motion smoothly over time as opposed to randomly sampling them  independently for each frame. This ensures that the augmented sequence still retains the temporal coherence  naturally present in videos.}
  \label{fig:temporal_augmentation}
\end{figure}

\subsection{Camera Motion Augmentation}

A crucial step in the synthetic video generation is camera motion augmentation (shown in Figure \ref{fig:temporal_augmentation}). Although it is not feasible to predict views of an arbitrarily moving camera without knowing the 3D structure, occluded parts and lighting sources in the scene, we can approximate it using affine image transformations. Here we consider the affine motion of a viewing frame over the video, which includes temporally smooth changes in rotation, translation, and scale. As we will show in section \ref{sec:experiments}, when we train without these augmentations, the training loss quickly decreases but the model does not transfer to real repetition videos. We empirically find camera motion augmentation is a vital part of training effective models with synthetic videos.

To achieve camera motion augmentations, we temporally vary the parameters for various motion types in a continuous manner as the video proceeds. For example, we change the angle of rotation smoothly over time. This ensures that the video is temporally coherent even after the augmentation. Figure \ref{fig:temporal_augmentation} illustrates how temporal augmentation parameter drives viewing frame (shown in blue rectangle) for each motion type. This results in videos with fewer near duplicates across the repeating segments. 
\section{\countix Dataset}
\label{sec:countix_dataset}

\begin{figure*}
\setkeys{Gin}{width=\linewidth} 
\renewcommand\thesubfigure{\thefigure\alph{subfigure}}
\begin{minipage}{0.74\textwidth}
    \includegraphics[height=4.6cm]{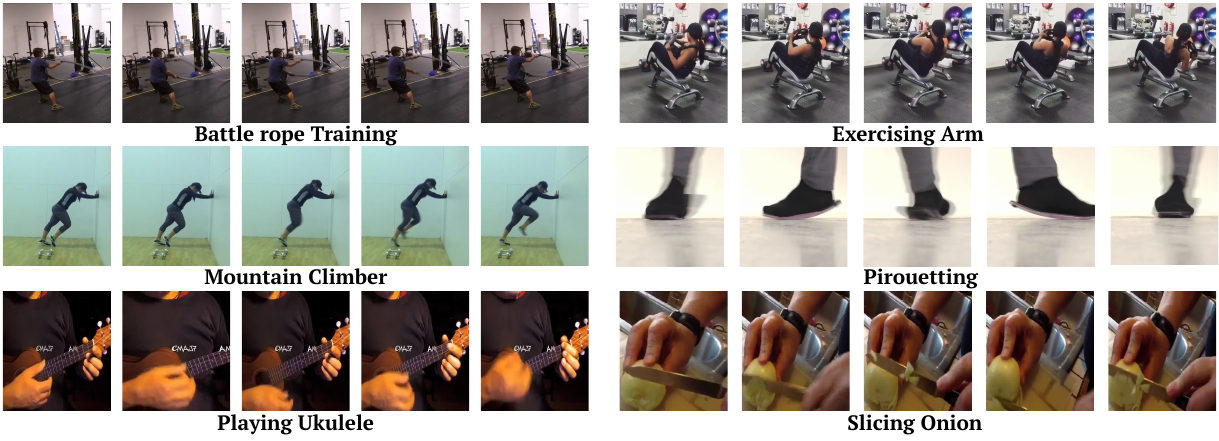}
\end{minipage}\hfill
\begin{minipage}{0.25\textwidth}
    \begin{subfigure}{\linewidth}
\includegraphics[height=2.3cm]{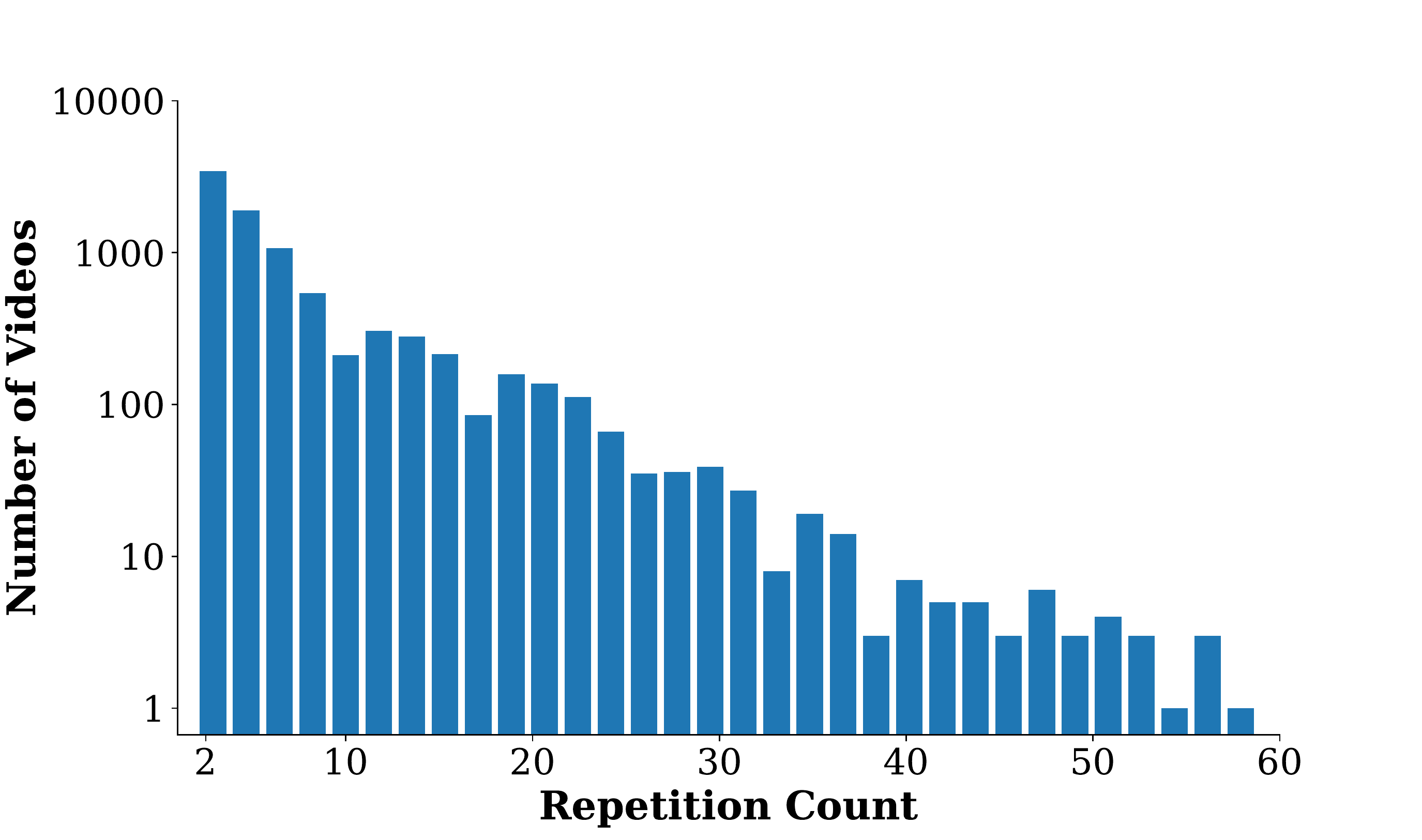}
    \end{subfigure}
    \begin{subfigure}{\linewidth}
    \includegraphics[height=2.3cm]{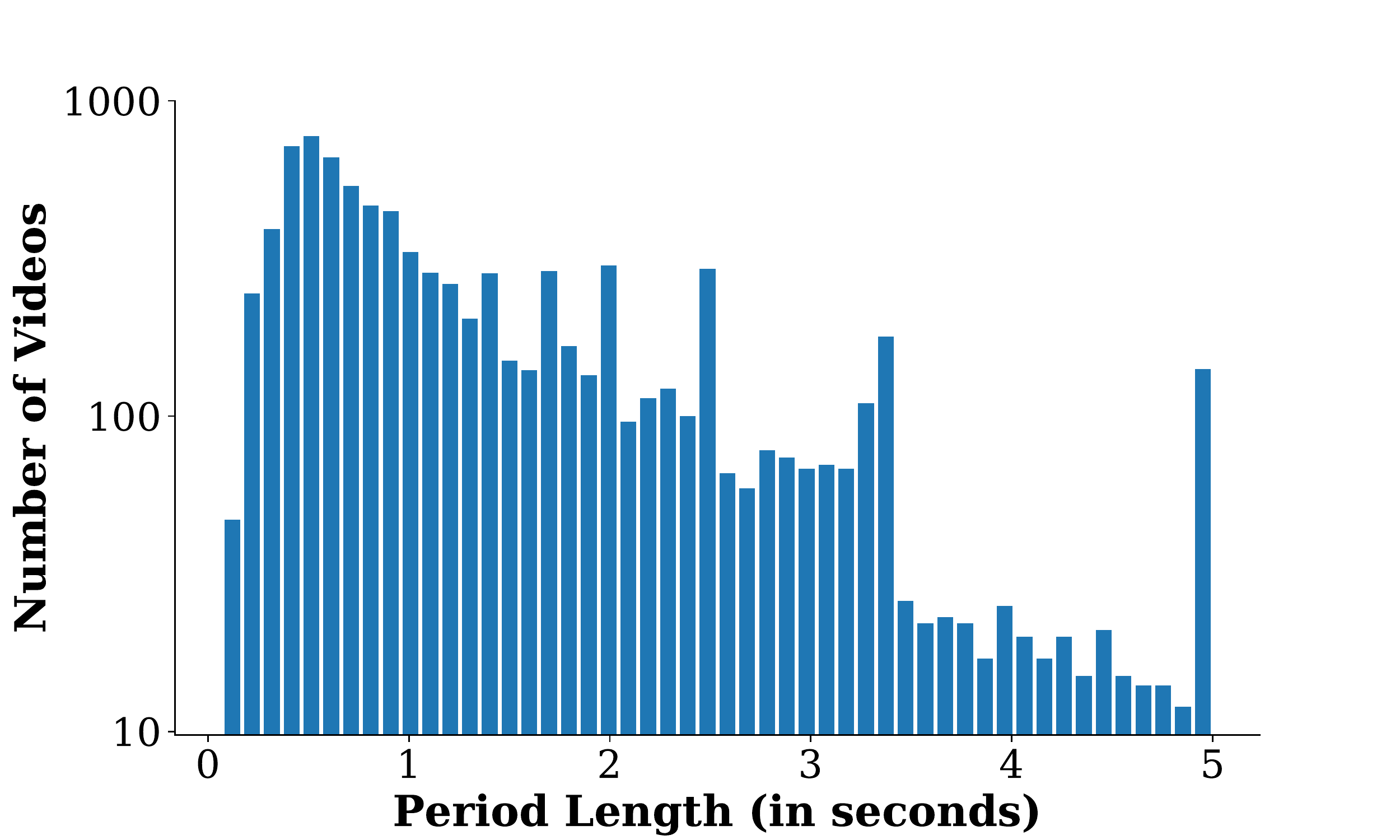}
    \end{subfigure}

\end{minipage}
\caption{\textbf{Countix dataset.} In the left two columns, we present examples of repeating videos from the Countix dataset. The last column shows the distribution of the number of the videos in the dataset with respect to the count and the period length labels.}
  \label{fig:countix_dataset}
\end{figure*}

Existing datasets for video repetition counting~\cite{runia2018real,levy2015live} are mostly utilized for testing purposes, mainly due to their limited size. The most recent and challenging benchmark on this task is the QUVA repetition dataset~\cite{runia2018real} which includes realistic repetition videos with occlusion, camera movement, and changes in speed of the repeated actions. It is composed of $100$ class-agnostic test videos, annotated with the count of repeated actions. Despite being challenging, its limited size makes it hard to cover diverse semantic categories of repetitions. Also training supervised deep models with this scale of data is not feasible.

To increase the semantic diversity and scale up the size of counting datasets, we introduce the \textbf{\countix} dataset: a real world dataset of repetition videos collected in the wild (i.e. YouTube) covering a wide range of semantic settings with significant challenges such as camera and object motion, diverse set of periods and counts, and changes in the speed of repeated actions. 

\countix include repeated videos of 
workout activities (squats, pull ups, battle rope training, exercising arm),
dance moves (pirouetting, pumping fist),
playing instruments (playing ukulele),
using tools repeatedly (hammer hitting objects, chainsaw cutting wood, slicing onion),
artistic performances (hula hooping, juggling soccer ball),
sports (playing ping pong and tennis)
and many others. Figure~\ref{fig:countix_dataset} illustrates some examples from the dataset as well as the distribution of repetition counts and period lengths. 

\noindent \textbf{Dataset Collection}: The \countix dataset is a subset of the Kinetics~\cite{kay2017kinetics} dataset annotated with segments of repeated actions and corresponding counts. During collection we first manually choose a subset of classes from Kinetics which have a higher chance of repetitions happening in them for e.g.\ \textit{jumping jacks}, \textit{slicing onion} etc., rather than classes like \textit{head stand} or \textit{alligator wrestling}. 

We crowd-source the labels for repetition segments and counts for the selected classes. The interface used is similar to what is typically used to mark out temporal segments for fine-grained action recognition\cite{gu2018ava, roth2019ava}. The annotators are asked to first segment the part of the video that contains valid repetitions with unambiguous counts. The annotators then proceed to count the number of repetitions in each segment. This count serves as the label for the entire clip. We reject segments with insignificant overlap in the temporal extents marked out by 3 different annotators. For the remaining segments, we use the median of the count annotations and segment extents as the ground truth.  The \textit{\countix} dataset is about 90 times bigger than the previous largest repetition counting dataset (QUVA Repetition Dataset). The detailed statistics can be viewed in Table~\ref{tab:dataset_stats}. The dataset is available on the project webpage.

Note that we retain the train/val/test splits from the Kinetics dataset. Hence, models pre-trained with Kinetics may be used for training counting models without data leakage.

\begin{table}[!h]
\setlength{\tabcolsep}{0.3em}
\centering
\footnotesize{
    \begin{tabular}{l|c|c}
    \toprule
     & \textbf{QUVA} & \textbf{\countix} \\
    \midrule
    No. of Videos in Train set & 0 & 4588\\
    No. of Videos in Val. set & 0 & 1450\\
    No. of Videos in Test set & 100 & 2719\\
    \midrule
    Duration Avg. $\pm$ Std (s) & 17.6 $\pm$ 13.3 & 6.13 $\pm$ 3.08 \\
    Duration Min./Max. (s) & 2.5 / 64.2 & 0.2 / 10.0 \\
    Count Avg $\pm$ Std & 12.5 $\pm$ 10.4& 6.84 $\pm$ 6.76 \\
    Count Min./Max. & 4 / 63 & 2 / 73\\
    \bottomrule
    \end{tabular}
\caption{Statistics of \countix and QUVA Repetition datasets.}
\label{tab:dataset_stats}
\vspace{-3em}
}    
\end{table}
\section{Experiments}
\label{sec:experiments}

We start by explaining the existing benchmarks and the evaluation metrics used in repetition counting. We next present a series of ablation studies that demonstrate which components and design choices are crucial. Then we compare our performance on the existing benchmarks and show that \repnet clearly outperforms the state-of-the-art methods on repetition counting and periodicity detection. Finally, through qualitative analysis, we bring more insight into our model.

\subsection{Benchmarks and Evaluation Metrics}
\label{ssec:benchmarks}

Here we discuss two established benchmark datasets for periodicity detection and repetition counting together with the commonly used evaluation metrics. 

\noindent \textbf{Periodicity detection}: The benchmark dataset for this task is the PERTUBE dataset~\cite{panagiotakis2018unsupervised}, which has per frame labels identifying periodicity, if the frame is a part of a repeating action or not. \cite{panagiotakis2018unsupervised} casts the problem as a binary per frame classification task and reports precision, recall, F1 score  and  overlap. We follow the same metrics for evaluation.

\noindent \textbf{Repetition counting}: 
As discussed in Section \ref{sec:countix_dataset}, the QUVA dataset~\cite{runia2018real} is the largest available dataset for repetition counting. 
The existing literature uses two main metrics for evaluating repetition counting in videos:

\noindent\textbf{Off-By-One (OBO) count error}. If the predicted count is within one count of the ground truth value, then the video is considered to be classified correctly, otherwise it is a mis-classification. The OBO error is the mis-classification rate over the entire dataset. 

\noindent\textbf{Mean Absolute Error (MAE) of count}. This metric measures the absolute difference between the ground truth count and the predicted count, and then normalizes it by dividing with the ground truth count. The reported MAE error is the mean of the normalized absolute differences over the entire dataset.

Both in our ablation experiments and state-of-the-art comparisons we follow \cite{runia2018real,levy2015live} and report OBO and MAE errors over the QUVA and \countix validation set. We also provide a final score on the \countix test set in Table~\ref{tab:countix_results}.

\subsection{Implementation Details}
We implement our method in Tensorflow~\cite{abadi2016tensorflow}. We initialize the encoder with weights from an ImageNet pre-trained ResNet-50 checkpoint. We train the model for $400K$ steps with a learning rate of $6 \times 10^{-6}$ with the ADAM optimizer and batch size of 5 videos (each with 64 frames). For all ablation studies we train the model on the synthetic repetition data unless otherwise stated. Additional details are provided on the project webpage. 

\subsection{Ablations}
We perform a number of ablations to justify the decisions made while designing \repnet. 
 
\noindent\textbf{Temporal Self-similarity Matrix (TSM):}
In Table~\ref{tab:data_ablation} we compare the impact of adding the TSM to the model. Models without the TSM apply the transformer directly on the per-frame embeddings produced by the encoder. The temporal self-similarity matrix substantially improves performance on all metrics and validation datasets whether we train the model using synthetic repetition videos, real \countix videos or a mix of both. Moreover, the TSM layer helps in generalizing to real repetition videos even when the model has only seen synthetic repetition videos (rows 1 and 2 in Table~\ref{tab:data_ablation}).

\noindent\textbf{Training Data Source:}
We vary the training data sources in Table~\ref{tab:data_ablation} while comparing our synthetic repetition videos with real ones from the \countix dataset. We find that \repnet achieves similar performance on the \countix  dataset when trained with synthetic videos or with the real repetition videos of the \countix dataset. But the model trained on \countix dataset is worse on the QUVA dataset compared to training on synthetic repeating videos. This shows using a synthetic repeating dataset results in a model that performs competitively on unseen classes as well. The best performance in terms of OBO error is achieved when the model is trained with both the datasets.

\noindent{\textbf{Alternative Period Prediction Architectures:}} In Table~\ref{tab:arch_ablation}, we compare the transformer architecture with other contemporary sequence models like LSTM and Temporal CNNs. We also compare it with a model that uses a 2D CNN on the self-similarity matrix itself. We find that the transformer architecture performs better than these alternatives.

\noindent{\textbf{Camera Motion Augmentation:}} In Table~\ref{tab:camera_motion_ablation} we show the value of camera motion augmentation when using the synthetic repeating dataset. We observe that performance on both datasets improves when the fraction of samples in the batch with camera motion augmentation is increased.

\begin{table}[!h]
\setlength{\tabcolsep}{0.3em}
\centering
\footnotesize{
    \begin{tabular}{c|c|c|c|c|c}
    \toprule
      \multicolumn{1}{c|}{} & &\multicolumn{2}{c|}{\textbf{QUVA}} & \multicolumn{2}{c}{\textbf{\countix (Val)}} \\
      \midrule
         \textbf{TSM} & \textbf{Training Data Source} & \textbf{MAE} &  \textbf{OBO} & \textbf{MAE} &  \textbf{OBO} \\
        \midrule
                &  Synthetic & 1.2853 & 0.64 & 1.1671 & 0.5510\\
          \checkmark & Synthetic &\textbf{ 0.1035} &\textbf{ 0.17} & \textbf{0.3100} & \textbf{0.2903}\\
        \midrule
                &  \countix & 0.7584 & 0.72 & 0.6483 & 0.5448\\
          \checkmark & \countix &  \textbf{0.3225} & \textbf{0.34} & \textbf{0.3468} & \textbf{0.2949}\\
          \midrule
           &  Synthetic + \countix & 0.6388 & 0.57 & 0.8889 & 0.4848\\
          \checkmark & Synthetic + \countix & \textbf{0.1315} & \textbf{0.15} &\textbf{0.3280} & \textbf{0.2752}\\
    \bottomrule
    \end{tabular}
\caption{Ablation of architecture with or without the temporal self-similarity matrix (TSM) with different training data sources.}
\label{tab:data_ablation}
}
\vspace{-3em}
\end{table}

\begin{table}[!h]
\setlength{\tabcolsep}{0.3em}
\centering
\footnotesize{
    \begin{tabular}{l|c|c|c|c}
    \toprule
      \multicolumn{1}{c|}{} & \multicolumn{2}{c|}{\textbf{QUVA}} & \multicolumn{2}{c}{\textbf{\countix (Val)}} \\
      \midrule
         \textbf{~~~Architecture} & \textbf{MAE} &  \textbf{OBO} & \textbf{MAE} &  \textbf{OBO} \\
         \midrule
         Transformer  & \textbf{0.1035} & \textbf{0.17} & \textbf{0.3100} & \textbf{0.2903}\\
         LSTM~\cite{hochreiter1997long} & 0.1395 & 0.18 & 0.6895 & 0.3579\\
         2D CNN & 0.1186 & \textbf{0.17} & 0.4440 & 0.3310\\
         1D Temporal CNN & 0.3229 & 0.23 & 0.7077 & 0.3641\\
    \bottomrule
    \end{tabular}
\caption{Performance of different period prediction architectures when trained with synthetic data.}
\label{tab:arch_ablation}
}\vspace{-3em}
\end{table}

\begin{table}[!h]
\setlength{\tabcolsep}{0.3em}
\centering
\footnotesize{
    \begin{tabular}{c|c|c|c|c}
    \toprule
      \multicolumn{1}{c|}{} & \multicolumn{2}{c|}{\textbf{QUVA}} & \multicolumn{2}{c}{\textbf{\countix (Val)}} \\
      \midrule
         \textbf{Augmentation Fraction} & \textbf{MAE} &  \textbf{OBO} & \textbf{MAE} &  \textbf{OBO} \\
         \midrule
         0.00 & 0.7178 & 0.32 & 1.2629 & 0.4683\\
         0.25 & 0.1414 & 0.17 & 0.4430 & 0.3303\\
         0.50 & 0.1202 & \textbf{0.15} & 0.3729 & 0.2993\\
         0.75 & \textbf{0.1035} & 0.17 & \textbf{0.3100} & 0.2903\\
         1.00 & 0.1710 & 0.17 & 0.3346 & \textbf{0.2848}\\
    \bottomrule
    \end{tabular}
\caption{Impact of camera motion augmentation when trained with synthetic data.}
\label{tab:camera_motion_ablation}
}\vspace{-3em}
\end{table}

\subsection{Evaluation on Benchmarks}

We compare our system with the current state-of-the-art methods on periodicity detection and repetition counting on the established benchmarks described in Section~\ref{ssec:benchmarks}. 

\noindent\textbf{Periodicity Detection.}
We report the performance for measuring periodicity classification by choosing the threshold that maximizes the F1 score. As done in \cite{panagiotakis2018unsupervised} we calculate the metrics on a per video basis and average the scores. We also report Area Under the Curve (AUC) of the precision-recall curve which is independent of the threshold chosen. Our model produces an AUC of 0.969. We outperform the previous work without using any hand-designed filtering methods mentioned in \cite{panagiotakis2018unsupervised} (see Table~\ref{tab:pertube_results}). Our model trained entirely on synthetic data works out of the box for the task of periodicity detection in real videos.

\noindent\textbf{Repetition Counting.}
In Table~\ref{tab:quva_results} we compare our \repnet model with previous models and show it outperforms existing methods by a significant margin and therefore establishing a new state-of-the-art for this dataset. Experimental results on the test set of  \countix dataset indicate that \repnet is an effective baseline for the video repetition counting task (see Table~\ref{tab:countix_results}).

\begin{table}[!h]
\setlength{\tabcolsep}{0.3em}
\centering
\footnotesize{
    \begin{tabular}{l|c|c|c|c}
    \toprule
       \textbf{\hspace{1.5cm}Model} & \textbf{Recall} & \textbf{Precision} & \textbf{F1} & \textbf{Overlap}\\
        \midrule
        Power spectrum baseline \cite{panagiotakis2018unsupervised} & 0.793 & 0.611 & 0.668 & 0.573\\
        P-MUCOS \cite{panagiotakis2018unsupervised} & 0.841 & 0.757 & 0.77 & 0.677\\
        \repnet (Ours) & \textbf{0.859} & \textbf{0.821} & \textbf{0.820} &  \textbf{0.731}\\
    \bottomrule
    \end{tabular}
\caption{Periodicity detection results on the PERTUBE Dataset}
\label{tab:pertube_results}
\vspace{-1em}
}
\end{table}

\begin{table}[!h]
\centering
\normalsize{
    \begin{tabular}{l|c|c}
    \toprule
       \textbf{\hspace{1.5cm} Model} & \textbf{MAE} & \textbf{OBO}\\
        \midrule
        Visual quasi-periodicity \cite{pogalin2008visual} & 0.385 & 0.51\\
        Live Repetition Counting \cite{levy2015live} & 0.482 & 0.55\\
        Div-Grad-Curl \cite{runia2018real} & 0.232 & 0.38\\
        \repnet (Ours) & \textbf{0.104} & \textbf{0.17}\\
    \bottomrule
    \end{tabular}
\caption{Counting Results on the QUVA dataset.}
\label{tab:quva_results}
\vspace{-1em}
}
\end{table}

\begin{table}[!h]
\centering
\normalsize{
    \begin{tabular}{c|c|c}
    \toprule
       \textbf{Model} & \textbf{MAE} & \textbf{OBO}\\
        \midrule
        \repnet & 0.3641 & 0.3034\\
    \bottomrule
    \end{tabular}
\caption{Counting Results on the \countix test set.}
\label{tab:countix_results}
\vspace{-1em}
}
\end{table}

\begin{figure}
  \includegraphics[width=0.48\textwidth]{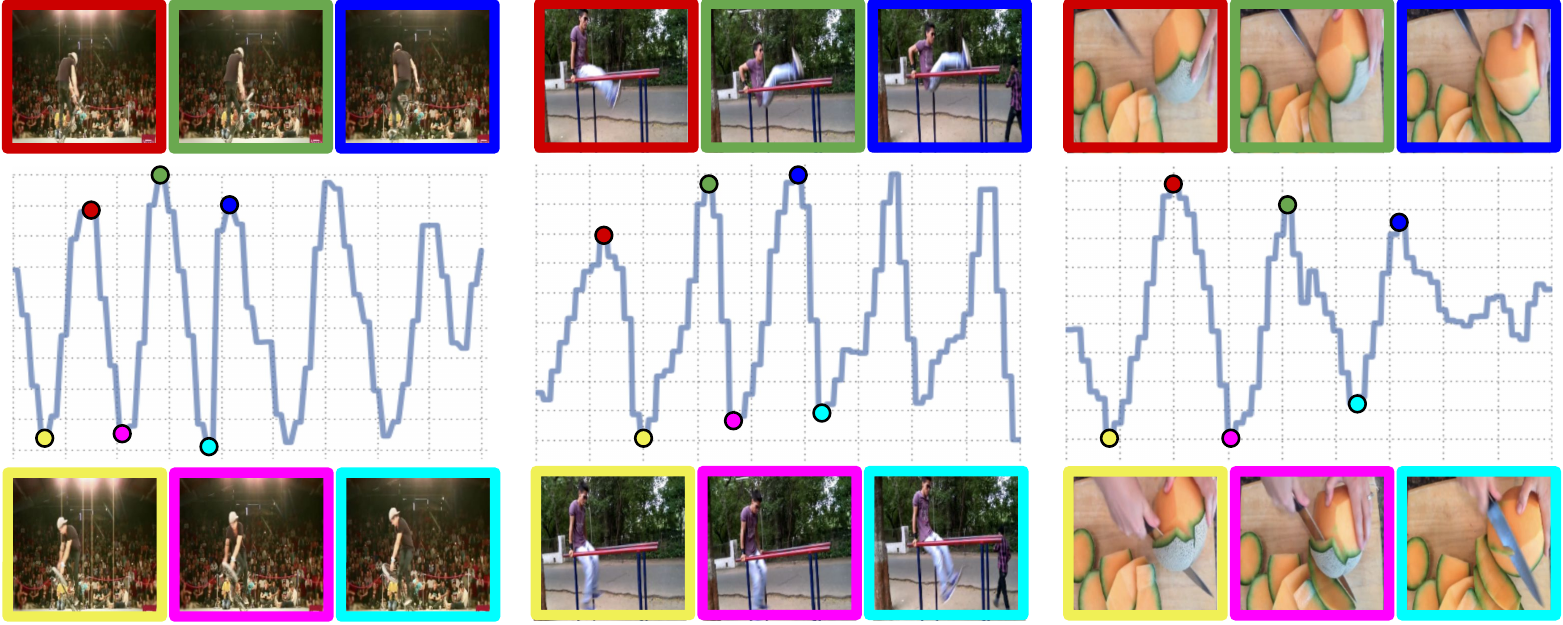}
  \caption{\textbf{1D PCA projections of the encoder features over time}. Note that even 1D projections of the learned features are encoding the periodicity fairly well. Frames with similar embeddings across different periods show similar states in the video (angle of rotation of biker, position of legs of person and position of knife). Best viewed with zoom. Video version on webpage \href{https://sites.google.com/view/repnet/1d-pca}{here}.}
  \label{fig:pca_projections}
\end{figure}

\begin{figure*}
  \includegraphics[width=\textwidth]{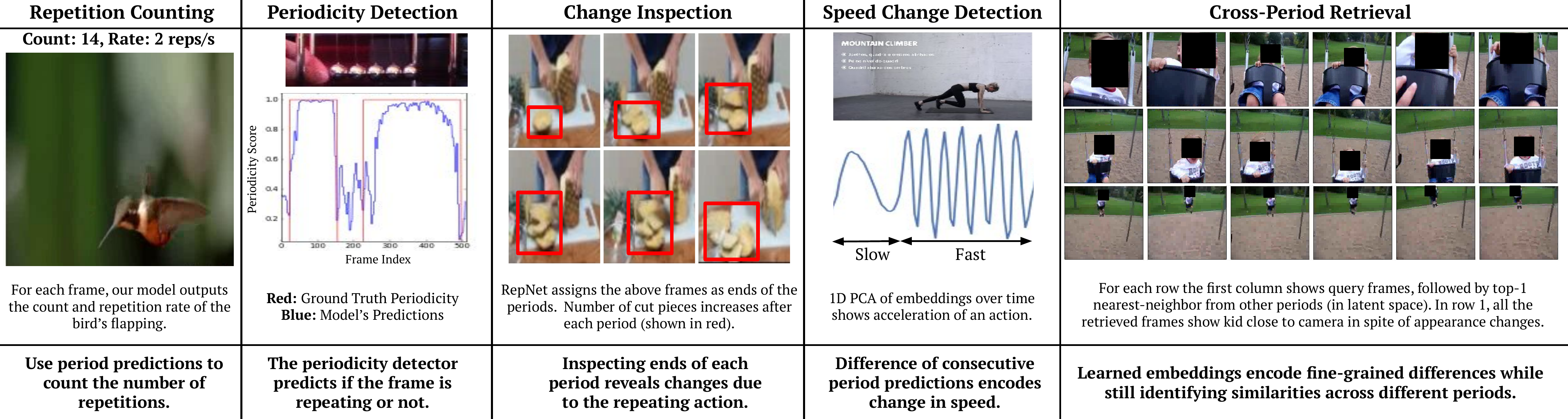}
  
  \caption{\textbf{One model, many domains and applications.} A single model is capable of performing these tasks over videos from many diverse domains (animal movement, physics experiments, humans manipulating objects, people exercising, child swinging) in a class-agnostic manner. Please see the \href{https://sites.google.com/view/repnet}{project webpage} for videos showcasing these tasks.}
  \label{fig:applications}
\end{figure*}

\subsection{Qualitative analysis}
\noindent\textbf{Temporal Self-similarity Matrix}. TSM provides us with meaningful interpretations about the model's predictions. It also contains additional information regarding acceleration and deceleration of the action. We show some examples of self-similarity matrices in Figure~\ref{fig:self_similarity_matrices}.

\noindent\textbf{1D PCA Embeddings}. We also investigate the learned embeddings which are used to produce the TSM. In Figure~\ref{fig:pca_projections}, we project the 512 dimensional vector to 1 dimension using the first principal component of the per-frame embeddings for each video. This reveals interesting quasi-sinusoidal patterns traced out by the embeddings in time. We plot the frames when the embeddings are changing directions and observe that the retrieved frames show the person or object in a similar state but in different periods.

\noindent\textbf{Double Counting Errors}. We observe that a common failure mode of our model is that for some actions (e.g. juggling soccer ball), it predicts half the count reported by annotators. This happens when the model considers left and right legs' motion for counting while people tend to consider the ball's up/down motion resulting in people double counting the repetitions. We believe such errors are difficult to isolate in a class-agnostic manner. But they can be fixed easily with either labeled data or post-processing methods if the application is known.

\section{Applications}

\noindent\textbf{Predict speed changes of repetitions}.
Our method takes in a video clip and predicts the period of any repeated action. The consecutive difference of predicted rates encodes the rate of speed change of the repetitions. Monitoring speed changes is useful for exercise tracking applications where it might be important to know if someone is speeding up or slowing down (Column 4 in Figure~\ref{fig:applications}). 

\noindent\textbf{Estimating frequency of processes from videos}.
Our model can be used to predict the count and frequency of repeating phenomena from videos for e.g.\ biological processes (heartbeats). \cite{wu2012eulerian} presented a method to reveal subtle changes by magnifying the difference in frames. We find that the output from the above system can be fed directly into our model to predict the frequency of these changes. A class-agnostic period estimator removes the need to explicitly train on these videos. On our project webpage, we show examples of repetition counting on echo-cardiogram videos which look very different from Kinetics videos .

\noindent\textbf{Fine-grained cross-period retrieval}.
The learned embeddings are useful for performing cross-period retrieval. In other words, the features capture similarities present across different periods while still encoding subtle differences between similar looking frames. Examples of these retrievals are shown in Figure \ref{fig:pca_projections} and the last column in Figure~\ref{fig:applications}.

\noindent\textbf{Repetitions with longer temporal extent}. Many repeating phenomena occur over a longer temporal scale (in the order of days or years). Even though our model has been trained on short videos ($\sim$10s), it can still work on videos with slow periodic events by automatically choosing a higher input frame stride. On the project webpage, we show videos where \repnet predicts the period length of a day from videos of the earth captured by satellites.

\noindent\textbf{Aid self-supervised video representation learning}.
Self-supervised learning methods for video embeddings, e.g. Shuffle and Learn~\cite{misra2016shuffle}, Odd-One-Out networks~\cite{fernando2017self}, DPC~\cite{han2019video}, TCC~\cite{dwibedi2019temporal} and TCN~\cite{sermanet2018time} are not designed to handle repetitions in sequences. \repnet can identify the repeating sections and may help in training on videos with repetitions without modifying the proposed objectives.
\section{Conclusion}
We have shown a simple combination of synthetic training data, together with an architecture using temporal self-similarity, results in a powerful class-agnostic repetition counting model. This model successfully detects periodicity and predicts counts over a diverse set of \textit{actors} (objects, humans, animals, the earth) and sensors (standard camera, ultrasound, laser microscope) and has been evaluated on a vast collection of videos. With this we have addressed the case of simple repetitions, and the next step is to consider more complex cases such as multiple simultaneous repeating signals and temporal arrangements of repeating sections such as in dance steps and music. 

\noindent \textbf{Acknowledgements:} We thank Aishwarya Gomatam, Anelia Angelova, Meghana Thotakuri, Relja Arandjelovic,  Shefali Umrania,  Sourish Chaudhuri, and Vincent Vanhoucke for their help with this project.

{\small
\bibliographystyle{ieee_fullname}
\bibliography{egbib}

\begin{thebibliography}{10}\itemsep=-1pt

\bibitem{abadi2016tensorflow}
Mart{\'\i}n Abadi, Paul Barham, Jianmin Chen, Zhifeng Chen, Andy Davis, Jeffrey
  Dean, Matthieu Devin, Sanjay Ghemawat, Geoffrey Irving, Michael Isard, et~al.
\newblock Tensorflow: A system for large-scale machine learning.
\newblock In {\em 12th $\{$USENIX$\}$ Symposium on Operating Systems Design and
  Implementation ($\{$OSDI$\}$ 16)}, pages 265--283, 2016.

\bibitem{alayrac2019visual}
Jean-Baptiste Alayrac, Joao Carreira, and Andrew Zisserman.
\newblock The visual centrifuge: Model-free layered video representations.
\newblock In {\em Proceedings of the IEEE Conference on Computer Vision and
  Pattern Recognition}, pages 2457--2466, 2019.

\bibitem{arteta2016counting}
Carlos Arteta, Victor Lempitsky, and Andrew Zisserman.
\newblock Counting in the wild.
\newblock In {\em European conference on computer vision}, pages 483--498.
  Springer, 2016.

\bibitem{belongie2004structure}
Serge Belongie and Josh Wills.
\newblock Structure from periodic motion.
\newblock In {\em International Workshop on Spatial Coherence for Visual Motion
  Analysis}, pages 16--24. Springer, 2004.

\bibitem{benabdelkader2001eigengait}
Chiraz BenAbdelkader, Ross Cutler, Harsh Nanda, and Larry Davis.
\newblock Eigengait: Motion-based recognition of people using image
  self-similarity.
\newblock In {\em International Conference on Audio-and Video-Based Biometric
  Person Authentication}, pages 284--294. Springer, 2001.

\bibitem{benabdelkader2004gait}
Chiraz BenAbdelkader, Ross~G Cutler, and Larry~S Davis.
\newblock Gait recognition using image self-similarity.
\newblock {\em EURASIP Journal on Advances in Signal Processing},
  2004(4):721765, 2004.

\bibitem{boominathan2016crowdnet}
Lokesh Boominathan, Srinivas~SS Kruthiventi, and R~Venkatesh Babu.
\newblock Crowdnet: A deep convolutional network for dense crowd counting.
\newblock In {\em Proceedings of the 24th ACM international conference on
  Multimedia}, pages 640--644. ACM, 2016.

\bibitem{chao2018rethinking}
Yu-Wei Chao, Sudheendra Vijayanarasimhan, Bryan Seybold, David~A Ross, Jia
  Deng, and Rahul Sukthankar.
\newblock Rethinking the faster r-cnn architecture for temporal action
  localization.
\newblock In {\em Proceedings of the IEEE Conference on Computer Vision and
  Pattern Recognition}, pages 1130--1139, 2018.

\bibitem{cutler2000robust}
Ross Cutler and Larry~S. Davis.
\newblock Robust real-time periodic motion detection, analysis, and
  applications.
\newblock {\em IEEE Transactions on Pattern Analysis and Machine Intelligence},
  22(8):781--796, 2000.

\bibitem{damen2018scaling}
Dima Damen, Hazel Doughty, Giovanni Maria~Farinella, Sanja Fidler, Antonino
  Furnari, Evangelos Kazakos, Davide Moltisanti, Jonathan Munro, Toby Perrett,
  Will Price, et~al.
\newblock Scaling egocentric vision: The epic-kitchens dataset.
\newblock In {\em Proceedings of the European Conference on Computer Vision
  (ECCV)}, pages 720--736, 2018.

\bibitem{dwibedi2019temporal}
Debidatta Dwibedi, Yusuf Aytar, Jonathan Tompson, Pierre Sermanet, and Andrew
  Zisserman.
\newblock Temporal cycle-consistency learning.
\newblock In {\em Proceedings of the IEEE Conference on Computer Vision and
  Pattern Recognition}, pages 1801--1810, 2019.

\bibitem{dwibedi2017cut}
Debidatta Dwibedi, Ishan Misra, and Martial Hebert.
\newblock Cut, paste and learn: Surprisingly easy synthesis for instance
  detection.
\newblock In {\em Proceedings of the IEEE International Conference on Computer
  Vision}, pages 1301--1310, 2017.

\bibitem{dwibedi2018learning}
Debidatta Dwibedi, Jonathan Tompson, Corey Lynch, and Pierre Sermanet.
\newblock Learning actionable representations from visual observations.
\newblock In {\em 2018 IEEE/RSJ International Conference on Intelligent Robots
  and Systems (IROS)}, pages 1577--1584. IEEE, 2018.

\bibitem{fernando2017self}
Basura Fernando, Hakan Bilen, Efstratios Gavves, and Stephen Gould.
\newblock Self-supervised video representation learning with odd-one-out
  networks.
\newblock In {\em Proceedings of the IEEE conference on computer vision and
  pattern recognition}, pages 3636--3645, 2017.

\bibitem{georgakis2017synthesizing}
Georgios Georgakis, Arsalan Mousavian, Alexander~C Berg, and Jana Kosecka.
\newblock Synthesizing training data for object detection in indoor scenes.
\newblock {\em arXiv preprint arXiv:1702.07836}, 2017.

\bibitem{gu2018ava}
Chunhui Gu, Chen Sun, David~A Ross, Carl Vondrick, Caroline Pantofaru, Yeqing
  Li, Sudheendra Vijayanarasimhan, George Toderici, Susanna Ricco, Rahul
  Sukthankar, et~al.
\newblock Ava: A video dataset of spatio-temporally localized atomic visual
  actions.
\newblock In {\em Proceedings of the IEEE Conference on Computer Vision and
  Pattern Recognition}, pages 6047--6056, 2018.

\bibitem{han2019video}
Tengda Han, Weidi Xie, and Andrew Zisserman.
\newblock Video representation learning by dense predictive coding.
\newblock In {\em Proceedings of the IEEE International Conference on Computer
  Vision Workshops}, pages 0--0, 2019.

\bibitem{hashimura2019collective}
H Hashimura, YV Morimoto, M Yasui, and M Ueda.
\newblock Collective cell migration of dictyostelium without camp oscillations
  at multicellular stages.
\newblock {\em Communications biology}, 2:34--34, 2019.

\bibitem{he2016deep}
Kaiming He, Xiangyu Zhang, Shaoqing Ren, and Jian Sun.
\newblock Deep residual learning for image recognition.
\newblock In {\em Proceedings of the IEEE conference on computer vision and
  pattern recognition}, pages 770--778, 2016.

\bibitem{hochreiter1997long}
Sepp Hochreiter and J{\"u}rgen Schmidhuber.
\newblock Long short-term memory.
\newblock {\em Neural computation}, 9(8):1735--1780, 1997.

\bibitem{junejo2010view}
Imran~N Junejo, Emilie Dexter, Ivan Laptev, and Patrick Perez.
\newblock View-independent action recognition from temporal self-similarities.
\newblock {\em IEEE transactions on pattern analysis and machine intelligence},
  33(1):172--185, 2010.

\bibitem{karvounas2019reactnet}
Giorgos Karvounas, Iason Oikonomidis, and Antonis Argyros.
\newblock Reactnet: Temporal localization of repetitive activities in
  real-world videos.
\newblock {\em arXiv preprint arXiv:1910.06096}, 2019.

\bibitem{kay2017kinetics}
Will Kay, Joao Carreira, Karen Simonyan, Brian Zhang, Chloe Hillier, Sudheendra
  Vijayanarasimhan, Fabio Viola, Tim Green, Trevor Back, Paul Natsev, Mustafa
  Suleyman, and Andrew Zisserman.
\newblock The kinetics human action video dataset.
\newblock {\em arXiv preprint arXiv:1705.06950}, 2017.

\bibitem{korner2013temporal}
Marco K{\"o}rner and Joachim Denzler.
\newblock Temporal self-similarity for appearance-based action recognition in
  multi-view setups.
\newblock In {\em International Conference on Computer Analysis of Images and
  Patterns}, pages 163--171. Springer, 2013.

\bibitem{kuehne2014language}
Hilde Kuehne, Ali Arslan, and Thomas Serre.
\newblock The language of actions: Recovering the syntax and semantics of
  goal-directed human activities.
\newblock In {\em Proceedings of the IEEE conference on computer vision and
  pattern recognition}, pages 780--787, 2014.

\bibitem{lempitsky2010learning}
Victor Lempitsky and Andrew Zisserman.
\newblock Learning to count objects in images.
\newblock In {\em Advances in neural information processing systems}, pages
  1324--1332, 2010.

\bibitem{levy2015live}
Ofir Levy and Lior Wolf.
\newblock Live repetition counting.
\newblock In {\em Proceedings of the IEEE International Conference on Computer
  Vision}, pages 3020--3028, 2015.

\bibitem{li2013automatic}
Wen Li and Dezhen Song.
\newblock Automatic bird species detection using periodicity of salient
  extremities.
\newblock In {\em 2013 IEEE International Conference on Robotics and
  Automation}, pages 5775--5780. IEEE, 2013.

\bibitem{li2018structure}
Xiu Li, Hongdong Li, Hanbyul Joo, Yebin Liu, and Yaser Sheikh.
\newblock Structure from recurrent motion: From rigidity to recurrency.
\newblock In {\em Proceedings of the IEEE Conference on Computer Vision and
  Pattern Recognition}, pages 3032--3040, 2018.

\bibitem{lu2018class}
Erika Lu, Weidi Xie, and Andrew Zisserman.
\newblock Class-agnostic counting.
\newblock In {\em Asian Conference on Computer Vision}, pages 669--684.
  Springer, 2018.

\bibitem{misra2016shuffle}
Ishan Misra, C~Lawrence Zitnick, and Martial Hebert.
\newblock Shuffle and learn: unsupervised learning using temporal order
  verification.
\newblock In {\em European Conference on Computer Vision}, pages 527--544.
  Springer, 2016.

\bibitem{niyogi1994analyzing}
Sourabh~A Niyogi and Edward~H Adelson.
\newblock Analyzing gait with spatiotemporal surfaces.
\newblock In {\em Proceedings of 1994 IEEE Workshop on Motion of Non-rigid and
  Articulated Objects}, pages 64--69. IEEE, 1994.

\bibitem{panagiotakis2018unsupervised}
Costas Panagiotakis, Giorgos Karvounas, and Antonis Argyros.
\newblock Unsupervised detection of periodic segments in videos.
\newblock In {\em 2018 25th IEEE International Conference on Image Processing
  (ICIP)}, pages 923--927. IEEE, 2018.

\bibitem{pintea2018hand}
Silvia~L Pintea, Jian Zheng, Xilin Li, Paulina~JM Bank, Jacobus~J van Hilten,
  and Jan~C van Gemert.
\newblock Hand-tremor frequency estimation in videos.
\newblock In {\em Proceedings of the European Conference on Computer Vision
  (ECCV)}, pages 0--0, 2018.

\bibitem{pogalin2008visual}
Erik Pogalin, Arnold~WM Smeulders, and Andrew~HC Thean.
\newblock Visual quasi-periodicity.
\newblock In {\em 2008 IEEE Conference on Computer Vision and Pattern
  Recognition}, pages 1--8. IEEE, 2008.

\bibitem{roth2019ava}
Joseph Roth, Sourish Chaudhuri, Ondrej Klejch, Radhika Marvin, Andrew
  Gallagher, Liat Kaver, Sharadh Ramaswamy, Arkadiusz Stopczynski, Cordelia
  Schmid, Zhonghua Xi, et~al.
\newblock Ava-activespeaker: An audio-visual dataset for active speaker
  detection.
\newblock {\em arXiv preprint arXiv:1901.01342}, 2019.

\bibitem{runia2018real}
Tom~FH Runia, Cees~GM Snoek, and Arnold~WM Smeulders.
\newblock Real-world repetition estimation by div, grad and curl.
\newblock In {\em Proceedings of the IEEE Conference on Computer Vision and
  Pattern Recognition}, pages 9009--9017, 2018.

\bibitem{seitz1997view}
Steven~M Seitz and Charles~R Dyer.
\newblock View-invariant analysis of cyclic motion.
\newblock {\em International Journal of Computer Vision}, 25(3):231--251, 1997.

\bibitem{sermanet2018time}
Pierre Sermanet, Corey Lynch, Yevgen Chebotar, Jasmine Hsu, Eric Jang, Stefan
  Schaal, Sergey Levine, and Google Brain.
\newblock Time-contrastive networks: Self-supervised learning from video.
\newblock In {\em 2018 IEEE International Conference on Robotics and Automation
  (ICRA)}, pages 1134--1141. IEEE, 2018.

\bibitem{SelfSim_ShechtmanIrani07}
Eli Shechtman and Michal Irani.
\newblock Matching local self-similarities across images and videos.
\newblock In {\em IEEE Conference on Computer Vision and Pattern Recognition
  2007 (CVPR'07)}, June 2007.

\bibitem{shou2016temporal}
Zheng Shou, Dongang Wang, and Shih-Fu Chang.
\newblock Temporal action localization in untrimmed videos via multi-stage
  cnns.
\newblock In {\em Proceedings of the IEEE Conference on Computer Vision and
  Pattern Recognition}, pages 1049--1058, 2016.

\bibitem{sigurdsson2016hollywood}
Gunnar~A Sigurdsson, G{\"u}l Varol, Xiaolong Wang, Ali Farhadi, Ivan Laptev,
  and Abhinav Gupta.
\newblock Hollywood in homes: Crowdsourcing data collection for activity
  understanding.
\newblock In {\em European Conference on Computer Vision}, pages 510--526.
  Springer, 2016.

\bibitem{stoica2005spectral}
Petre Stoica, Randolph~L Moses, et~al.
\newblock Spectral analysis of signals.

\bibitem{sun2015exploring}
Chuan Sun, Imran~Nazir Junejo, Marshall Tappen, and Hassan Foroosh.
\newblock Exploring sparseness and self-similarity for action recognition.
\newblock {\em IEEE Transactions on Image Processing}, 24(8):2488--2501, 2015.

\bibitem{tremblay2018training}
Jonathan Tremblay, Aayush Prakash, David Acuna, Mark Brophy, Varun Jampani, Cem
  Anil, Thang To, Eric Cameracci, Shaad Boochoon, and Stan Birchfield.
\newblock Training deep networks with synthetic data: Bridging the reality gap
  by domain randomization.
\newblock In {\em Proceedings of the IEEE Conference on Computer Vision and
  Pattern Recognition Workshops}, pages 969--977, 2018.

\bibitem{Varol_2017_CVPR}
Gul Varol, Javier Romero, Xavier Martin, Naureen Mahmood, Michael~J. Black,
  Ivan Laptev, and Cordelia Schmid.
\newblock Learning from synthetic humans.
\newblock In {\em The IEEE Conference on Computer Vision and Pattern
  Recognition (CVPR)}, July 2017.

\bibitem{vaswani2017attention}
Ashish Vaswani, Noam Shazeer, Niki Parmar, Jakob Uszkoreit, Llion Jones,
  Aidan~N Gomez, {\L}ukasz Kaiser, and Illia Polosukhin.
\newblock Attention is all you need.
\newblock In {\em Advances in neural information processing systems}, pages
  5998--6008, 2017.

\bibitem{vlachos2005periodicity}
Michail Vlachos, Philip Yu, and Vittorio Castelli.
\newblock On periodicity detection and structural periodic similarity.
\newblock In {\em Proceedings of the 2005 SIAM international conference on data
  mining}, pages 449--460. SIAM, 2005.

\bibitem{wang2013action}
Heng Wang and Cordelia Schmid.
\newblock Action recognition with improved trajectories.
\newblock In {\em Proceedings of the IEEE international conference on computer
  vision}, pages 3551--3558, 2013.

\bibitem{wu2012eulerian}
Hao-Yu Wu, Michael Rubinstein, Eugene Shih, John Guttag, Fr{\'e}do Durand, and
  William Freeman.
\newblock Eulerian video magnification for revealing subtle changes in the
  world.
\newblock 2012.

\bibitem{xie2018rethinking}
Saining Xie, Chen Sun, Jonathan Huang, Zhuowen Tu, and Kevin Murphy.
\newblock Rethinking spatiotemporal feature learning: Speed-accuracy trade-offs
  in video classification.
\newblock In {\em Proceedings of the European Conference on Computer Vision
  (ECCV)}, pages 305--321, 2018.

\bibitem{xie2018microscopy}
Weidi Xie, J~Alison Noble, and Andrew Zisserman.
\newblock Microscopy cell counting and detection with fully convolutional
  regression networks.
\newblock {\em Computer methods in biomechanics and biomedical engineering:
  Imaging \& Visualization}, 6(3):283--292, 2018.

\bibitem{yun2019cutmix}
Sangdoo Yun, Dongyoon Han, Seong~Joon Oh, Sanghyuk Chun, Junsuk Choe, and
  Youngjoon Yoo.
\newblock Cutmix: Regularization strategy to train strong classifiers with
  localizable features.
\newblock {\em arXiv preprint arXiv:1905.04899}, 2019.

\bibitem{zhang2017mixup}
Hongyi Zhang, Moustapha Cisse, Yann~N Dauphin, and David Lopez-Paz.
\newblock mixup: Beyond empirical risk minimization.
\newblock {\em arXiv preprint arXiv:1710.09412}, 2017.

\end{thebibliography}
}

\clearpage
\appendix

\section*{Appendix}
On our \href{https://sites.google.com/view/repnet}{project webpage} we provide visualizations of qualitative results, dataset samples, and 1D PCA visualizations.

\section{Qualitative Results}

All examples below have been created with a single model trained only with synthetic data.

\subsection{Counting on Videos with Different Sensors}
We provide examples of our model on different sensors:

\noindent\textbf{Echocardiogram.} \repnet can estimate the heartbeat rate from echocardiogram videos. We find the predicted heartrates close to the true heart rate measured by the device itself (\href{https://sites.google.com/view/repnet/ecg}{link to videos}) . Note how the same model works across different ECG machines.

\noindent\textbf{Laser Microscope.} We found videos of repeating biological phenomena in \cite{hashimura2019collective} where they observed a cellular phenomena under the laser microscope which results in spiral patterns in the video. We find that our model works out of the box measuring the rate at which the spirals are rotating. The model also captures the speed change in the process being measured (\href{https://sites.google.com/view/repnet/microscope}{link to videos}).

\noindent\textbf{Eulerian Magnified Videos} Our model works on videos produced by using Eulerian magnification to highlight subtle changes in time~\cite{wu2012eulerian}. We show \repnet can count on those videos without further training (\href{https://sites.google.com/view/repnet/eulerian}{link to videos}).

\subsection{Physics Experiments with \repnet}
We show examples of 2 videos where pendulums of different lengths are swung. The ratio of time periods can be used to predict the ratio of lengths of the pendulums. Our model can replace the step in which the people conducting the experiment measure time period with a stopwatch. We conduct the experiment from the video ourselves and find the ratio of time periods of the long to short pendulum using the period lengths predicted by our model to be 1.566. Based on the physics equations of oscillations of pendulums, the expected approximate ratio of the periods using approximation of the length of pendulum (from pixels in the video) is 1.612. We provide details in the experiments (\href{https://sites.google.com/view/repnet/physics}{here}).

\subsection{Consistent Multi-view Period Predictions}
We test our model of different views capturing the collapse of the Tacoma Bridge in 1940 due to resonance. Our model recovers the frequency of repetition from different viewpoints robustly (\href{https://sites.google.com/view/repnet/bridge}{link to videos}). 

\subsection{Inspecting Changes over Periods}

\begin{figure*}
  \includegraphics[width=\textwidth]{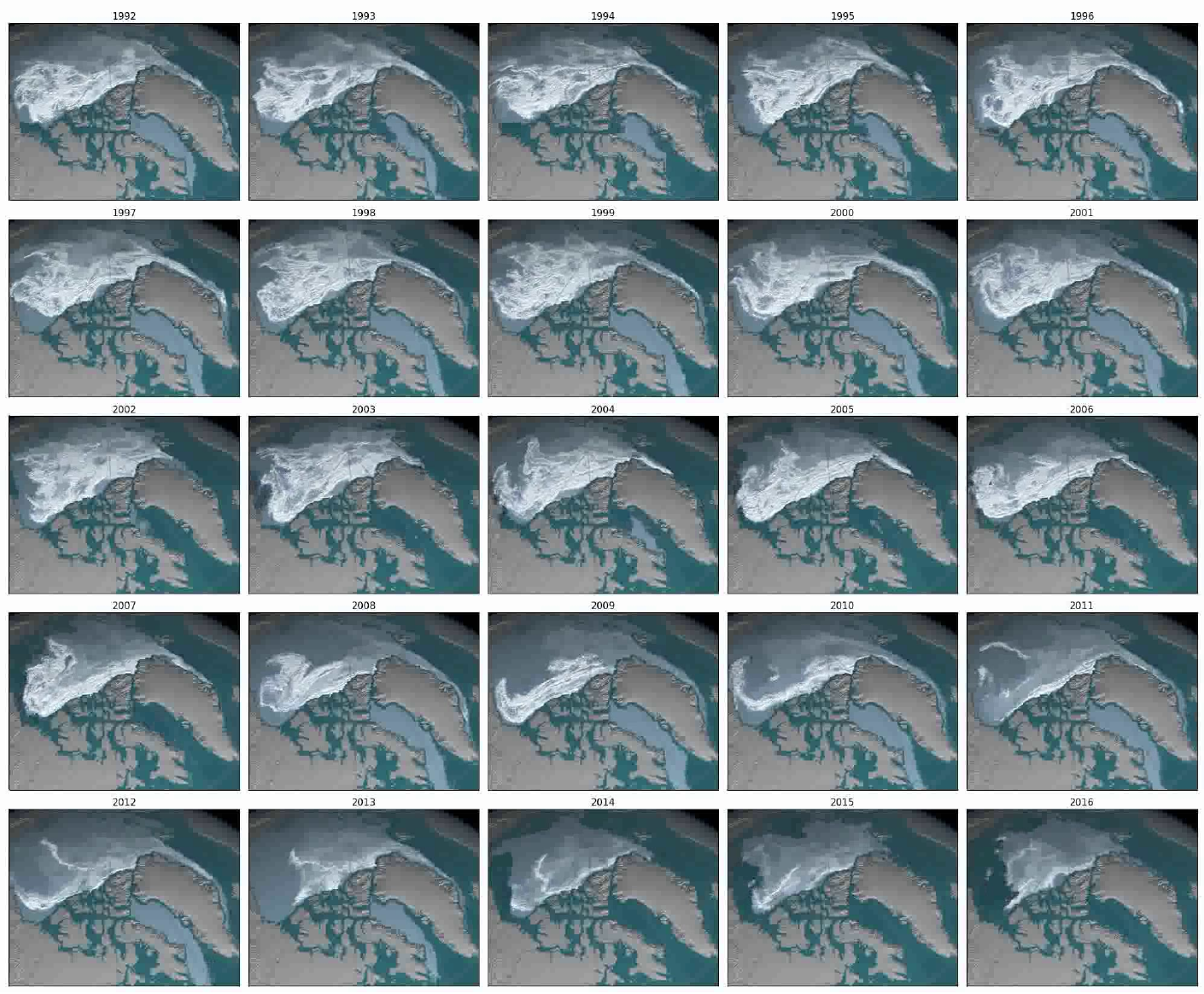}
  \caption{\textbf{Arctic Ice Cover Trends.} \repnet extracts frames that are each one period away showing decreasing ice cover over the years.}
  \label{fig:arctic_ice}

\end{figure*}

\repnet takes input of satellite image representation (released by NASA Goddard) of the ice-cover on the Arctic over the period of ~25 years and predicts the period to be roughly 1 year. In Figure~\ref{fig:arctic_ice}, we show frames that the model marks these frames one period (approximately a year) apart. The amount of ice cover visibly reduces over the years.

\subsection{Video Galleries}

We also provide video galleries with the following visualizations:
\begin{enumerate}[noitemsep,nolistsep]
    \item 1D PCA of the embeddings in time (\href{https://sites.google.com/view/repnet/1d-pca}{link to videos})
     \item Learned Temporal Self-similarity Matrices (TSMs) of different videos (\href{https://sites.google.com/view/repnet/tsms}{link to videos})
\end{enumerate}

\section{Ablations}
\subsection{Removing \countix Classes from Synthetic Data}
We show that removing the classes used in \countix from the pool of classes used for creating the synthetic data has marginal impact on performance (see Table~\ref{tab:remove_kinetics_ablation}). This shows that generalization of the \repnet model does not require the presence of \countix classes in the synthetic dataset highlighting the class-agnostic aspect of our model.

\begin{table}[!h]
\setlength{\tabcolsep}{0.3em}
\centering
\footnotesize{
    \begin{tabular}{l|c|c|c|c}
    \toprule
      \multicolumn{1}{l|}{} & \multicolumn{2}{c|}{\textbf{QUVA}} & \multicolumn{2}{c}{\textbf{\countix (Val)}} \\
      \midrule
      \textbf{Training Classes} & \textbf{MAE} &  \textbf{OBO} & \textbf{MAE} &  \textbf{OBO} \\
        \midrule
        Kinetics & \textbf{0.1035} & \textbf{0.17} & \textbf{0.3100} & \textbf{0.2903}\\
        Kinetics without \countix Classes & 0.1181 & 0.17 & 0.3751 & 0.2993\\
    \bottomrule
    \end{tabular}
\caption{Effect of removing \countix classes from synthetic training data is marginal.}
\label{tab:remove_kinetics_ablation}
}
\end{table}

\subsection{ImageNet Pre-training}
We evaluate the importance of ImageNet pretraining for the \repnet model and report the results in Table~\ref{tab:encoder_ablation}. We find that if we train the model completely from scratch (Row 2) we achieve performance that is ~4\% worse than with pre-training (Row 4). This performance still exceeds the current state of the art methods in repetition counting. Also, ImageNet initialization of the encoder without any further training (Row 3) is good enough for the repetition counting task due to the subsequent modules (TSM and transformer).

\begin{table}[!h]
\setlength{\tabcolsep}{0.3em}
\centering
\footnotesize{
    \begin{tabular}{c|c|c|c|c|c}
    \toprule
     & & \multicolumn{2}{c|}{\textbf{QUVA}} & \multicolumn{2}{c}{\textbf{\countix (Val)}} \\
     \midrule
      \textbf{Train base CNN} & \textbf{ImageNet Pre-trained} & \textbf{MAE} &  \textbf{OBO} & \textbf{MAE} &  \textbf{OBO} \\
        \midrule
         \xmark & \xmark & 0.3097 & 0.30 & 0.4928 & 0.3938\\
        \checkmark & \xmark & 0.1394 & 0.20 & 0.3877 & 0.3290\\
        \xmark & \checkmark  & 0.1270 & 0.19 & 0.3178 & 0.2910\\
        \checkmark & \checkmark & \textbf{0.1035 }&\textbf{0.17} & \textbf{0.3100} &\textbf{0.2903}\\
    \bottomrule
    \end{tabular}
\caption{Ablation of pre-training with ImageNet and training the base network or not. For all experiments,     we train the 3D conv and period prediction module.}
\label{tab:encoder_ablation}
}
\end{table}

\subsection{Camera Motion Augmentations}
We use various camera motion augmentation techniques to the synthetic repeating videos as described in Figure 5 in main paper. We show the effect of omitting different data augmentations. Each of these methods results in about 1.5\% to 2\%  worse OBO error and about 13\% to 26\% worse MAE error. Based on these experiments, we use all these augmentation techniques for rest of our experiments. 

\begin{table}[!h]
\setlength{\tabcolsep}{0.3em}
\centering
\footnotesize{
    \begin{tabular}{l|c|c|c|c}
    \toprule
      \multicolumn{1}{c|}{} & \multicolumn{2}{c|}{\textbf{QUVA}} & \multicolumn{2}{c}{\textbf{\countix (Val)}} \\
      \midrule
         \textbf{~~~Data Augmentation} & \textbf{MAE} &  \textbf{OBO} & \textbf{MAE} &  \textbf{OBO} \\
         \midrule
         With all augmentations  & \textbf{0.1035 }&\textbf{ 0.17} & \textbf{0.3100} &\textbf{ 0.2903}\\
        \midrule
          No scale &  0.1222 & 0.18 & 0.5751 & 0.3193\\
          No rotation &  0.1158 & 0.16 & 0.4406 & 0.3041\\
          No translation & 0.1202 & 0.16 & 0.5400 & 0.3069\\
          No reversed concatenation & 0.1211 & 0.16 & 0.4449 & 0.3131\\
    \bottomrule
    \end{tabular}

\caption{Effect of different camera motion augmentations when trained with synthetic data.}
\label{tab:data_aug_ablation}
}
\end{table}

\subsection{Varying Number of frames}
In Table~\ref{tab:num_frame_ablation}, we report results when we vary the number of frames which \repnet takes as input and find that $N=64$ frames provides us with the best performance. We use this setting for all the experiments in the main paper.

\begin{table}[!h]
\setlength{\tabcolsep}{0.3em}
\centering
\footnotesize{
    \begin{tabular}{c|c|c|c|c}
    \toprule
      \multicolumn{1}{c|}{} & \multicolumn{2}{c|}{\textbf{QUVA}} & \multicolumn{2}{c}{\textbf{\countix (Val)}} \\
      \midrule
         \textbf{Num Frames} & \textbf{MAE} &  \textbf{OBO} & \textbf{MAE} &  \textbf{OBO} \\
         \midrule
         32 & 0.1407 & 0.22 & 0.4800 & 0.3069\\
         64 & \textbf{0.1035 }& 0.17 & \textbf{0.3100} &\textbf{ 0.2903}\\
         96 & 0.1094 & \textbf{0.16} &	0.4870 & 0.3097\\
         128 & 0.1233 & 0.17 & 0.3429 & 0.3200\\
    \bottomrule
    \end{tabular}
\caption{Effect of varying number of frames in the clip.}
\label{tab:num_frame_ablation}
}
\end{table}

\subsection{Other Architectural Choices}
We also varied certain architectural choices made while designing \repnet but found they have minor impact on overall performance. 

\begin{table}[!h]
\setlength{\tabcolsep}{0.3em}
\centering
\footnotesize{
    \begin{tabular}{l|c|c|c|c}
    \toprule
      \multicolumn{1}{c|}{} & \multicolumn{2}{c|}{\textbf{QUVA}} & \multicolumn{2}{c}{\textbf{\countix (Val)}} \\
      \midrule
         \textbf{Architecture} & \textbf{MAE} &  \textbf{OBO} & \textbf{MAE} &  \textbf{OBO} \\
         \midrule
         Baseline  & \textbf{0.1035 }& 0.17 & \textbf{0.3100} & 0.2903\\
        \midrule
          No 3D Conv. &  0.1198 & \textbf{0.16} &	0.4478 & 0.3014\\
          No 2D Conv. before Transformer &  0.1586	& 0.19 & 0.5039 & 0.3048\\
          Replace L2 dist. with cosine sim. & 0.1153 & 0.18 & 0.3114 & 0.2972 \\
          No softmax & 0.1163 & \textbf{0.16} & 0.3835 & \textbf{0.2883}\\
    \bottomrule
    \end{tabular}

\caption{Effect of architecture variations when trained with synthetic data.}
\label{tab:minor_arch_ablation}
}
\end{table}

\section{Implementation Details}

\subsection{Detailed Architecture}
In Table~\ref{tab:archiecture} we present the detailed version of \repnet architecture.

\subsection{Architectures of Alternative Baselines}
 \textbf{2D CNN Baseline.} Our 2D CNN consists of the following convolutional layers $[32, 64, 128, 256, 512]$ each with filter size $3\times3$. After each convolution layer there is a max-pooling operation of $2\times2$ size with stride 2. Global spatial average pooling is done over the final feature map which is used to classify the period length of the entire clip. We also experimented with ResNet50 architecture and got similar performance.

  \textbf{LSTM.} We use the standard LSTM implemented in Tensorflow Keras library with 512 units.
  
\textbf{1D Temporal CNN.} We use 7 layers of temporal convolutions with dilation rates $[1,2,4,8,16,32,64]$. Each convolution layer is of size $512$ and has a kernel size of $2$ and has batch normalization. We use skip-connections with residuals for each layer.

\subsection{Combining Period Length and Periodicity Outputs during Inference}
Our model can be used to jointly detect periodic segments in the video and count repetitions only within the repeating segments. To do so, we sample consecutive windows of $N$ frames and provide it as input to \repnet which outputs per-frame periodicity $p_i$ and period lengths $l_i$. We define \textit{per-frame count} as $c_i = \frac{1}{l_i}$ if $p_i > T$ else 0, where T is a chosen threshold for the classifier. Count of the video is the sum of all per-frame counts: $\sum\limits_{i=1}^N c_i$.

\begin{table*}[hbt!]
\centering
\renewcommand{\arraystretch}{1.50}
\begin{tabular}{c|c|c|c}
\hline
\textbf{Module} & \textbf{Layer} & \textbf{Output Size} & \textbf{Layer Parameters/Notes} \\ \hline
\multirow{5}{*}{\textbf{Base Network}}  & conv1 & 56$\times$56$\times$64 & 7$\times$7, 64, stride 2\\
 \cline{2-4} & \multirow{2}{*}{conv2\_x} &  \multirow{2}{*}{28$\times$28$\times$256} & 3$\times$3 max pool, stride 2\\
\cline{4-4} & & & \blockb{64}{256}{3} \\
\cline{2-4} & conv3\_x & 14$\times$14$\times$512 & \blockb{128}{512}{4} \\
\cline{2-4} & conv4\_x & 7$\times$7$\times$1024 & \blockb{256}{1024}{3} \\
\hline
\multirow{2}{*}{\textbf{Temporal Context}} & Temporal Stacking & 64$\times$7$\times$7$\times$ 1024 & Stack features from all frames in time axis\\
\cline{2-4} & 3D Convolution & 64$\times$7$\times$7$\times$512 & $[3 \times 3 \times 3, 512]$, dilation rate $=$ 3\\
\hline
 \textbf{Dimensionality Reduction} & Spatial Pooling & 64$\times$512 & Global 2D Max-Pool\\
\hline
\multirow{3}{*}{\textbf{Temporal Self-similarity Matrix}} &  Pairwise L2 Distance & 64$\times$64 & \\
 \cline{2-4} &  Multiply with $-1$ & 64$\times$64 & Convert distances to similarities\\
 \cline{2-4} &  Row-wise Softmax & 64$\times$64 & Softmax temperature $=$ 13.5\\
 \hline
 \multirow{4}{*}{\textbf{Period Predictor}} & 2D Convolution & 64$\times$64$\times$32&  3$\times$3, 32\\
 \cline{2-4} & Transformer & 64$\times$64$\times$512 & 4 heads, 512 dims, learned positional embeddings\\
 \cline{2-4} & Flatten & 64$\times$32768 & Shared input for following 2 layers\\
 \cline{2-4} & Period Length Classifier & 64$\times$32 & \blockd{512}{512}{32}\\
 \cline{2-4} & Periodicity Classifier & 64$\times$1 & \blockd{512}{512}{1}\\
 \hline
\end{tabular}
\caption{\textbf{Detailed Architecture of \repnet.} The parameters in the form of: (1) $[n \times n, c]$ refers to 2D Convolution filter size and number of channels respectively (2) $[n \times n \times n, c]$ refers to 3D Convolution filter size and number of channels respectively (3) $[c]$ refers to channels in a fully-connected layers. }
\label{tab:archiecture}
\end{table*}

\section{Dataset Details}

\subsection{\countix Details}

The list of classes chosen for data collection while creating \countix dataset is mentioned in Table~\ref{tab:countix_classes}.

\begin{table*}[!h]
\setlength{\tabcolsep}{0.3em}
\centering
\footnotesize{
    \begin{tabular}{l|l}
    \toprule
   \textbf{ Kinetics Class Name} & \textbf{Description of the Repetitions}\\
    \midrule
battle rope training & number of times the person moves the battle ropes up to down\\
bench pressing & number of times the person lifts the bar to the top \\
bouncing ball (not juggling) & number of times has bounced the ball on the foot\\
bouncing on bouncy castle & number of times a person has jumped on the bouncy castle\\
bouncing on trampoline &  number of times a person has jumped on the trampoline\\
clapping & number of times someone claps\\
crawling baby & number of steps taken by baby\\
doing aerobics & number of times an aerobic step is repeated by the group or person\\
exercising arm & number of times the exercise is done by the person\\
front raises & number of times the weights are raised to the top in front of the person’s chest\\
gymnastics tumbling & number of times the gymnast completes a rotation\\
hammer throw & number of times the person rotates before throwing the hammer\\
headbanging & number of times have moved their head up and down\\
hula hooping & number of times the hula hoop moves about a person’s waist\\
juggling soccer ball & number of times the soccer ball is bounced\\
jumping jacks & number of times a person completes one step of jumping jack motion\\
lunge & number of times a person completes one step of lunge action\\
mountain climber (exercise) & number of times a person completes one step of mountain climber action\\
pirouetting & number of times the person rotates about their own axis\\
planing wood & number of times someone moves their hand back and forth while planing wood\\
playing ping pong & number of times the ball goes back and forth \\
playing tennis & number of times the ball goes back and forth\\
playing ukulele & number of times a hand strums up and down , count the number of times the hand reaches the top while strumming the guitar\\
pull ups & number of pull ups by counting the number of times a person reaches the top of the trajectory\\
pumping fist & number of times people move their fists\\
push up & number of pull ups by counting the number of times a person reaches the top of the trajectory \\
rope pushdown & number of times a person pulls down on the rope, count how many times they reach the bottom of the trajectory\\
running on treadmill & number of steps/strides taken by a person\\
sawing wood & number of times the saw goes back and forth \\
shaking head & number of times a person shakes their head\\
shoot dance & number of times a person completes a dance step\\
situp & number of times a person completes a situp motion, count the number of times the person reaches top of trajectory\\
skiing slalom & number of times a person bends to the side to change direction of velocity\\
skipping rope & number of times a person skips the rope, count the number of times the rope is at the top of trajectory\\
slicing onion & number of times the knife slices the onions\\
spinning poi & number of rotations completed by the lights\\
squat & number of times a person squats, count the number of times they reached the bottom of the trajectory\\
swimming butterfly stroke & number of times a person does a butterfly stroke\\
swimming front crawl & number of times a person does a front crawl\\
swinging on something & number of times a swing is completed, count the number of times the person is nearest to the camera\\
tapping pen & number of times a person taps the pen \\
triple jump & number of jumps done by person\\
using a wrench & number of times a wrench is rotated\\
using a sledge hammer & number of times a sledge hammer is brought down on an object, count the number of times the hammer hits the object\\

    \bottomrule
    \end{tabular}

\caption{Classes present in the \countix dataset along with descriptions of repetitions contained in them.}
\label{tab:countix_classes}
}
\end{table*}

\end{document}